\documentclass{article}
\usepackage{iclr2027_conference,times}
\usepackage[T1]{fontenc}
\usepackage[utf8]{inputenc}
\usepackage{amsmath,amssymb,booktabs,array,longtable,tabularx,multirow}
\usepackage{graphicx,float}
\usepackage{capt-of}
\usepackage{hyperref}
\usepackage{url}
\hypersetup{
  colorlinks=true,
  linkcolor=blue,
  citecolor=blue,
  urlcolor=blue,
  pdftitle={SyncRA: Learning Temporal Correspondence in Omni-Modal Models},
  pdfauthor={Zelong Xu, Yan Li, Wenhe Hu, Xiyang Hu}
}
\title{SyncRA: Learning Temporal Correspondence in Omni-Modal Models}
\author{
Zelong Xu$^{1}$\thanks{Corresponding author: \texttt{zxu684@wisc.edu}.}
\quad
Yan Li$^{1}$
\quad
Wenhe Hu$^{2}$
\quad
Xiyang Hu$^{3}$\\
$^{1}$University of Wisconsin--Madison \quad
$^{2}$University of Alberta \quad
$^{3}$Arizona State University
}

\iclrfinalcopy 
\begin{document}
\maketitle
\lhead{Preprint}
\begin{abstract}
Recent omni-modal models demonstrate strong perception of audio and visual inputs, yet often struggle to connect what they hear with what they see at the same moment. This weakness in temporal correspondence can cause models to associate spoken cues with the wrong visual scenes, producing plausible answers grounded in incorrect audio-visual pairings. We diagnose this problem through controlled temporal swaps, revealing that model answers do not reliably follow changes in these pairings. To address it, we propose Synchrony-Guided Representation Alignment (SyncRA), a lightweight method for strengthening temporal correspondence between audio and vision. Specifically, SyncRA contrasts intermediate audio--visual representations within each video, aligning matching moments while separating mismatched ones to capture local temporal correspondence within a shared global context. The objective derives supervision directly from existing input timing, requiring no additional annotations and leaving inference unchanged. We evaluate SyncRA across four open omni-modal models spanning different sizes and architectures on five public video benchmarks. SyncRA consistently outperforms answer-only fine-tuning across all model--benchmark combinations, while substantially improving the ability to track changing audio--visual pairings in controlled evaluations. These results demonstrate that lightweight, targeted supervision can effectively strengthen temporal correspondence, alongside consistent improvements in audio--visual question answering.
\end{abstract}
\section{Introduction}
\label{sec:introduction}

Omni-modal models that jointly understand text, audio, and video have advanced rapidly \citep{xu2025qwen25omni,cui2026minicpmo45,xu2025qwen3omni,nvidia2026nemotronomni}. By encoding interleaved sensory streams and reasoning over them in a shared context, these models achieve strong results across a wide range of video understanding tasks and are becoming general-purpose interfaces for real-world multimedia content. As they are applied to content that unfolds over time, a basic requirement is that they preserve the temporal relation between modalities: the model should know which visual observation accompanies a given sound or utterance, rather than only recognizing what appears somewhere in the same clip. We refer to this ability as local audio--visual temporal correspondence.

\begin{figure*}[t]
    \centering
    \includegraphics[width=1\textwidth]{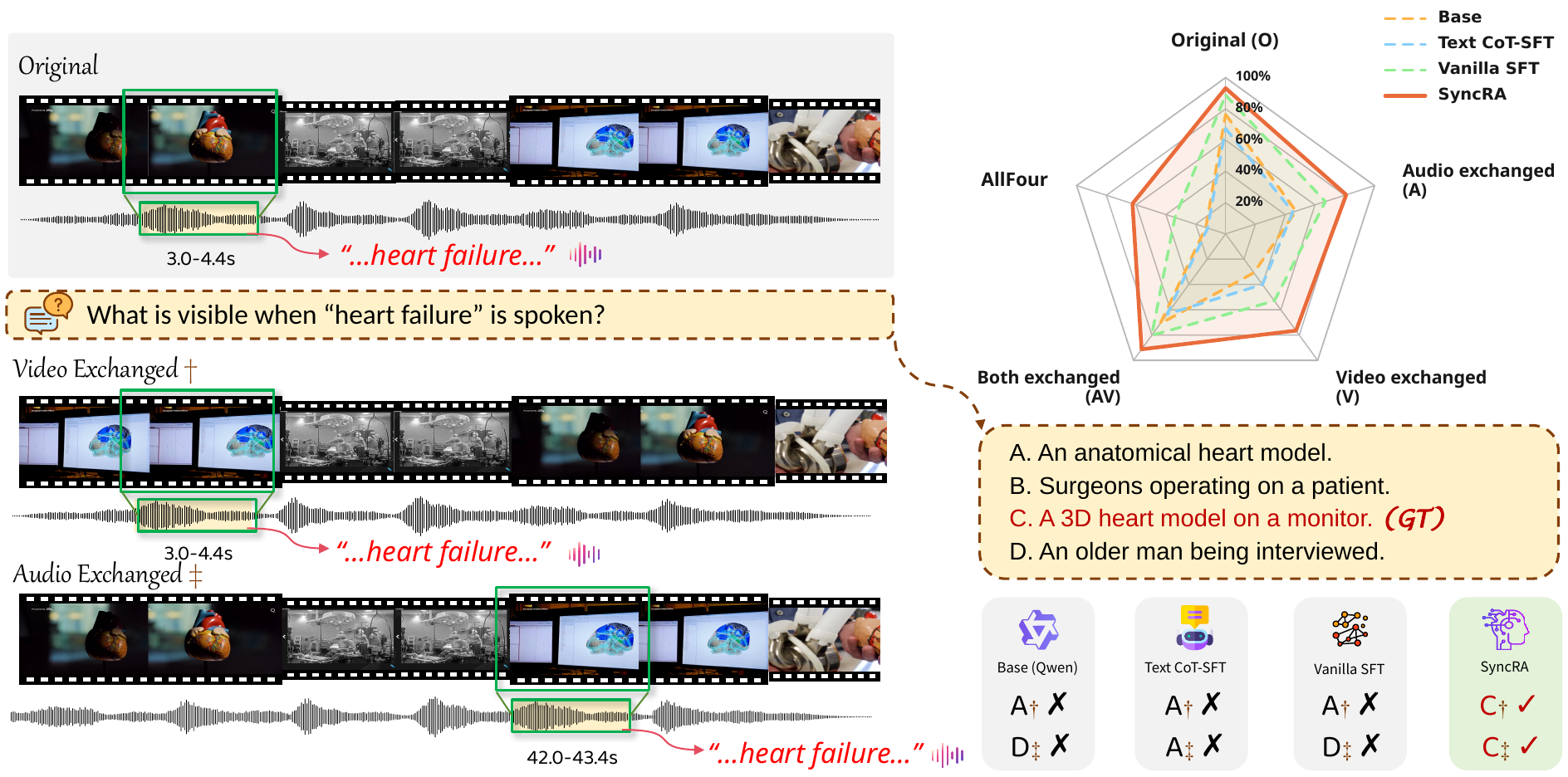}
    \caption{
        \textbf{A motivating example and the swap diagnostic.}
        \textbf{Left:} exchanging only the audio or only the video
        windows changes which scene is visible while ``heart
        failure'' is spoken (correct option A~$\to$~C).
        \textbf{Right:} Qwen3-Omni accuracy on the four input
        versions (top; AllFour credits a cue only if all four are
        answered correctly) and predictions on the two illustrated
        exchanges (bottom; $\dagger$/$\ddagger$). SyncRA answers both
        correctly. Fine-tuned checkpoints use Run~1.
    }
    \label{fig:correspondence-overview}
\end{figure*}

Failure of this correspondence can cause a model to attach a spoken cue to the wrong visual scene while still producing a fluent and plausible answer. Such errors are particularly difficult for users to detect because both the audio and visual content may individually be correct, while their temporal pairing is not. Figure~\ref{fig:correspondence-overview} illustrates such a case: while the narrator says ``heart failure'', an anatomical heart model is on screen; if the video windows are exchanged while the audio remains unchanged, the scene shown during that phrase changes, and a model that tracks the correspondence should change its answer accordingly.

This weakness is widespread rather than model-specific: independent evaluations consistently find that current omni-modal models fail at deep audio--visual temporal integration \citep{zhou2025dailyomni,chen2026avid,zhang2026avtrace}. Our diagnostic measures its local-pairing form: whether a model tracks which scene accompanies a given utterance. The problem is also structural: temporally aligned audio--visual data is scarce \citep{zhou2025dailyomni}, joint video--audio tokens account for only a small fraction of general-purpose pretraining \citep{xu2025qwen3omni}, and answer-level supervision leaves local pairing implicit, because a question answerable from global semantics or a single modality does not require distinguishing correct from incorrect temporal pairings.

To measure this failure directly, we construct a controlled swap diagnostic around spoken cues. For each source video, we identify four spoken cues that occur in four nonoverlapping temporal windows and ask which scene is visible while each cue is spoken. We then create three altered versions of the same video by exchanging the corresponding audio windows, the video windows, or both. Exchanging only one stream changes the correct answer because it changes which sound and scene occur together; exchanging both streams preserves the original pairing while moving it to another position. A model can therefore answer all versions correctly only if its prediction follows the audio--visual pairing present in the input, rather than either modality alone or the video's global semantics (Section~\ref{sec:gap}). Applied to four recent open omni-modal backbones, this diagnostic shows that current models often answer as if the original pairing were still in place, and that answer-only fine-tuning, the standard way such models are adapted to question answering, does not by itself repair this behavior. Strong accuracy on ordinary audio--visual QA therefore does not guarantee that a model tracks the temporal pairing itself.

Existing approaches leave temporal correspondence weak at two levels. The first is representational: encoding time within each modality does not ensure that audio and visual representations are aligned in time. Our probing analysis of a representative backbone finds that elapsed time remains linearly decodable from intermediate audio and visual states, yet direct matching of simultaneous audio and visual states is weak and deteriorates toward the final layers (Section~\ref{sec:representation}). Related work on audio--visual speech recognition similarly reports that deeper decoder representations become less sensitive to temporal order \citep{wang-etal-2026-bridging}. The model therefore retains temporal information without organizing the two modalities into a shared temporal structure. The second gap is supervisory: as discussed above, neither the training data nor the answer-level objective provides an explicit signal for local pairing. Together, these gaps call for making local temporal correspondence an explicit, representation-level target during ordinary QA fine-tuning, using a supervision signal that requires no additional annotation.

Building on audio--visual synchrony and temporal contrastive learning \citep{owens2018multisensory,araujo2025cavmaesync,dai2026latentomni}, we introduce Synchrony-Guided Representation Alignment (SyncRA), whose design mirrors these two gaps. To close the representation gap, SyncRA acts where the problem lives: one intermediate layer of the answering model. At this layer, it mean-pools the audio and visual hidden states within each interval, maps both modalities through a shared linear projection, and optimizes a symmetric InfoNCE objective that pulls simultaneous audio--visual states together while separating mismatched ones. To close the supervision gap, SyncRA derives the pairing signal from the input itself: the model's native processor already divides the audio and visual streams into shared temporal intervals, so observations from the same interval form positive pairs and other intervals of the same video serve as negatives, with no event-level annotation or external representation teacher. The auxiliary objective is optimized jointly with ordinary answer supervision, making local temporal correspondence an explicit representation-learning target. SyncRA is deliberately lightweight: the projection head adds only $64d$ training parameters and is discarded after training, leaving the architecture and inference cost unchanged. Because its interface relies only on token timing metadata rather than a specific attention layout, the same recipe applies across dense and mixture-of-experts backbones. SyncRA targets local input-time correspondence; narration referents, off-screen sound semantics, and desynchronization detection are out of scope.

We conduct controlled swap evaluations, broad benchmark comparisons, and representation analyses to test whether SyncRA strengthens temporal correspondence and whether the effect generalizes. On the swap diagnostic, SyncRA raises AllFour accuracy over answer-only fine-tuning on all four backbones, from 34.00\% to 62.50\% on Qwen3-Omni. On five public video understanding benchmarks, SyncRA improves mean accuracy in all 20 model--benchmark combinations. Auxiliary controls without correct pairing recover up to 73\% of this gain, so the pairing-specific claims rest on the swap diagnostic. Representation analyses link the learned alignment to the answers: in Qwen3-Omni's own unprojected states, same-interval retrieval at the final block rises from 3.44\% under answer-only fine-tuning to 31.06\% R@1, retrieval follows the current input position under an audio delay, and activation patching shows that the answer reads the cue-time states.

Our contributions are as follows. (1) We introduce SyncRA, a lightweight representation-level objective that directly supervises local audio--visual temporal correspondence during ordinary QA fine-tuning, without additional annotations or inference-time components. Across four dense and MoE backbones, SyncRA improves correspondence-sensitive answering and raises mean accuracy on all 20 model--benchmark combinations. (2) We develop a controlled swap test that directly measures whether a model's answers follow the audio--visual pairing present in its input, and show that this capability remains weak in current omni-modal models even after answer-only fine-tuning. (3) We identify the conditions under which classic synchrony supervision repairs pairing-following in omni-modal QA: same-video negatives, interval granularity, and the InfoNCE form. The resulting correspondence is directly visible in the backbone's own unprojected states.

\section{Methodology}
\label{sec:method}

\begin{figure*}[t]
    \centering
    \includegraphics[width=1\textwidth]{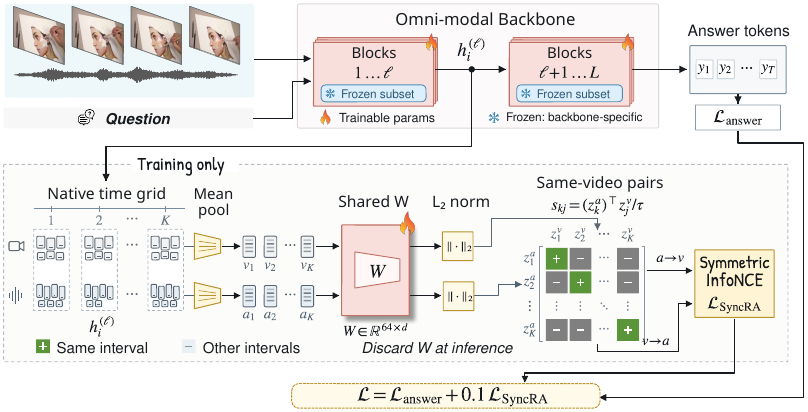}
    \caption{
        \textbf{Overview of Synchrony-Guided Representation
        Alignment (SyncRA).}
        \textbf{Top:} the ordinary QA path, trained with
        $\mathcal{L}_{\mathrm{answer}}$.
        \textbf{Bottom (training only):} at a selected block $\ell$,
        audio and visual states are mean-pooled within native
        temporal intervals, projected by a shared $W$ with $L_2$
        normalization, and contrasted with symmetric InfoNCE against
        other intervals of the same video.
    }
    \label{fig:syncra-overview}
\end{figure*}

Figure~\ref{fig:syncra-overview} summarizes SyncRA, which adds
local audio--visual correspondence supervision to intermediate
backbone states during ordinary QA fine-tuning.

\subsection{Temporal interface and interval pooling}

Given a video, its audio track, and a question, an omni-modal answering model encodes perceptual inputs and processes their tokens in a joint context before generating an answer. SyncRA operates on these existing hidden states at one selected block.

Let \(h_{i}^{(\ell)}\in\mathbb{R}^{d}\) denote the contextualized state of a perceptual token at the output of one-indexed block \(\ell\) in a joint audio--video forward pass. Audio and visual labels identify token origin. Native timestamps, interval boundaries, or temporal patch metadata assign each token to a shared set of intervals along the video timeline. For each interval $k$ containing both modalities, we mean-pool the corresponding token sets \(A_{k}\) and \(V_{k}\):
\begin{equation}\label{eq:pool}
a_k=\frac{1}{|A_k|}\sum_{i\in A_k}h_i^{(\ell)},\qquad v_k=\frac{1}{|V_k|}\sum_{i\in V_k}h_i^{(\ell)}.
\end{equation}
The valid intervals form a set of size $K$. These intervals are processor sampling units, not annotated semantic events. A same-interval pair is positive because the observations co-occur in the input; they need not describe the same object or event. Membership follows time metadata rather than token adjacency, so processors that place modality tokens in separate sequence blocks are also supported. Appendix~\ref{app:media} specifies each backbone's grid.

\subsection{Within-video synchrony contrast}

A shared bias-free projection \(W\in\mathbb{R}^{64\times d}\) maps both modalities from hidden dimension $d$ to 64 dimensions. We normalize the projected states and use cosine similarity with temperature \(\tau\):
\begin{equation}\label{eq:similarity}
z_k^a=\frac{Wa_k}{\|Wa_k\|_2},\qquad z_k^v=\frac{Wv_k}{\|Wv_k\|_2},\qquad s_{kj}=\frac{(z_k^a)^\top z_j^v}{\tau}.
\end{equation}
The per-video auxiliary objective classifies the corresponding interval in both directions:
\begin{equation}\label{eq:syncraloss}
\mathcal L_{\mathrm{SyncRA}}=-\frac{1}{2K}\sum_{k=1}^K\left[\log\frac{\exp(s_{kk})}{\sum_{j=1}^K\exp(s_{kj})}+\log\frac{\exp(s_{kk})}{\sum_{j=1}^K\exp(s_{jk})}\right].
\end{equation}
All candidates come from the same video. This choice holds source identity and global context fixed, requiring discrimination among times within a recording. It follows the broader use of synchrony as supervision \citep{korbar2018cooperative,owens2018multisensory}. 

\subsection{Joint training with answer supervision}

We add the auxiliary objective to final-answer token cross-entropy:
\begin{equation}\label{eq:joint}
\mathcal L=\mathcal L_{\mathrm{answer}}+\lambda\mathcal L_{\mathrm{SyncRA}},\qquad \lambda=0.1,\quad \tau=0.07.
\end{equation}

We fix $\tau=0.07$ following standard contrastive practice \citep{he2020momentum} and set $\lambda=0.1$; Section~\ref{sec:loss_ablation} shows that the gains hold across $\lambda$.

Samples with fewer than two valid intervals receive answer supervision alone. Gradients update W and the permitted backbone parameters; answer supervision also trains subsequent layers. The auxiliary head adds only $64d$ training parameters and no inference-time component, since $W$ is discarded after training.

We fix one supervision block per backbone using shared linear probes fitted on frozen-checkpoint states before SyncRA training. The selected one-indexed blocks are 21/28 for Qwen2.5-Omni, 9/36 for MiniCPM-o 4.5, 36/48 for Qwen3-Omni, and 26/52 for Nemotron. QA fine-tuning initializes a new $W$. Section~\ref{sec:layer_analysis} explains the rationale; Appendix~\ref{app:layer_selection} specifies the selection procedure.

\section{Experimental Setup}
\label{sec:setup}

\subsection{Data}

We use a fixed, balanced 10K-example subset of OmniVideo-100K \citep{cai2026omnivideo100k} for all fine-tuning experiments. It contains 10,000 questions from 1,782 videos, with 5,000 multiple-choice and 5,000 open-ended questions across ten task categories. Clips last 60--180 seconds. The source-disjoint split comprises 9,500 training questions from 1,367 videos and 500 validation questions from 415 videos. All methods share this split and the same media inputs. Working with a compact, category-balanced corpus is a deliberate choice: SyncRA's supervision comes from input timing rather than from additional data or annotation, and since Vanilla SFT is trained on the same data with the same budget, any improvement is attributable to the supervision signal rather than to data scale. The timing labels are obtained for free from the processor and scale to larger corpora without annotation cost (Appendix~\ref{app:discussion}). Appendix~\ref{app:training} provides the data and processing details.

\subsection{Backbones and baselines}

To evaluate portability across architectures, we use two dense backbones, Qwen2.5-Omni-7B and MiniCPM-o 4.5, and two mixture-of-experts (MoE) backbones, Qwen3-Omni-30B-A3B and Nemotron-3-Nano-Omni-30B-A3B \citep{xu2025qwen25omni,cui2026minicpmo45,xu2025qwen3omni,nvidia2026nemotronomni}. Dense models undergo full understanding fine-tuning, while MoE models use partial parameter updates with routed experts frozen. The update scope is held fixed across training objectives within each backbone.

Our main comparisons include \emph{Base}, the released checkpoint without task-specific fine-tuning; \emph{Vanilla SFT} (answer-only fine-tuning), which supervises only final answers after removing the supplied reasoning text; and \emph{Text CoT-SFT}, which supervises the supplied reasoning followed by the answer. SyncRA uses the same final-answer targets as Vanilla SFT and adds the local correspondence objective in Section~\ref{sec:method}. Auxiliary-objective variants are introduced in the ablation study.

\subsection{Training configuration}

All fine-tuned methods start from the corresponding released checkpoint and
share one optimization recipe (AdamW, learning rate $10^{-5}$, effective
batch size 12, two epochs); SyncRA keeps the auxiliary configuration of
Section~\ref{sec:method}. Checkpoints are selected by the lowest
validation answer-token negative log-likelihood, never by benchmark scores
or post-training representation diagnostics; Text CoT-SFT is selected
on its complete reasoning-and-answer targets. Full hyperparameters,
trainable modules, layer-selection procedures, and Text CoT-SFT settings
appear in Appendices~\ref{app:optimization}, \ref{app:layer_selection},
and~\ref{app:text_cot}.

\subsection{Evaluation}

We evaluate WorldSense (3,172 questions), Daily-Omni (1,197), OmniVideoBench (1,000), LVOmniBench (1,014), and AVUT-Human (AVUT's AV-Human split; 1,733 after excluding one item with an empty correct option), covering general, long-context, and audio-centered video understanding \citep{hong2026worldsense,zhou2025dailyomni,li2025omnivideobench,tao2026lvomnibench,yang2025avut}. We report accuracy on each benchmark and their equally weighted mean (Avg-5). Vanilla SFT and SyncRA each use three independent training runs, summarized by the mean and sample standard deviation; Base and Text CoT-SFT each use a single checkpoint.

Within each backbone, all methods use identical evaluation questions, media, and denominators. Evaluation uses greedy decoding and fixed answer extraction, with generation and parsing failures counted as incorrect. Appendices~\ref{app:benchmark_protocol} and~\ref{app:text_cot} specify prompts, media processing, and parsing rules.

\section{A Controlled Swap Test for Temporal Pairing}
\label{sec:gap}

To test whether a model's answers follow the current audio--visual pairing, we design a swap test around spoken cues. For each cue we ask: ``While the phrase \{cue\} is being spoken, which of the following scenes is visible?'' The four options describe scenes present in the source clip. We use 100 source videos (81 from Daily-Omni and 19 from LVOmniBench), disjoint from all training and validation sources. Each video contributes four unique spoken cues in four equal-length, nonoverlapping windows, with a stable, unique visible scene during each utterance. Independent ASR and human inspection establish the annotations before model inference.

Each cue is evaluated independently on original (O), audio-exchanged (A), video-exchanged (V), and jointly exchanged (AV) inputs. We swap windows $1\leftrightarrow3$ and $2\leftrightarrow4$ in the designated streams, preserving within-window order, content outside the windows, and total duration. A single-stream exchange changes the correct scene, while joint exchange preserves the original pairing at a new position. Questions and option order stay the same across versions; neither timestamps nor version labels are given to the model.

Each checkpoint answers 400 questions per version and 1,600 in total. The primary metric, AllFour, credits a cue only when all four versions are answered correctly; overall accuracy averages all 1,600 question--version evaluations. AllFour shares the joint-success motivation of MMBench's CircularEval and Winoground's group score, but varies temporal pairing rather than option order or image--text composition \citep{liu2024mmbench,thrush2022winoground}. Paired 95\% bootstrap intervals resample source videos with all their cues and versions, with checkpoints held fixed. Appendix~\ref{app:behavior} provides the construction and scoring details.

\section{Results}
\label{sec:results}

We evaluate whether SyncRA improves answers that depend on the current audio--visual pairing and whether its benefits extend across architectures and benchmark tasks.

\subsection{Correspondence improves across architectures}
\label{sec:behavior}

SyncRA improves correspondence-sensitive answering on both dense and both MoE backbones: on the same diagnostic of 100 source videos, it raises AllFour accuracy over Vanilla SFT for every backbone (Table~\ref{tab:behavior}). Gains range from 5.00 to 28.50 percentage points, and all four paired 95\% source-video bootstrap intervals exclude zero (Table~\ref{tab:app_cross_backbone_correspondence_deltas}).

For every backbone, the largest version-specific improvement occurs under a single-stream exchange, where the correct answer must follow a changed audio--visual pairing. Complete condition-level and Text CoT-SFT results appear in Table~\ref{tab:correspondence-full}; prediction matrices in Appendix~\ref{app:prediction_matrices} illustrate how answers follow the current pairing.

\begin{table}[t]
\centering
\small
\renewcommand{\arraystretch}{1.12}
\setlength{\tabcolsep}{9pt}
\caption{
\textbf{Correspondence-sensitive answering improves across architectures.}
AllFour accuracy (\%) on the same 100 source videos.
$\Delta$ is SyncRA minus Vanilla SFT in percentage points.
All four paired 95\% source-video bootstrap intervals exclude zero
(Appendix~\ref{app:behavior_metrics}).
}
\label{tab:behavior}
\begin{tabular}{@{}lcrrrr@{}}
\toprule
Backbone & Arch. & Base & Vanilla SFT & SyncRA & $\Delta$ \\
\midrule
Qwen2.5-Omni  & Dense & 2.25  & 11.00 & \textbf{20.75} & \textbf{+9.75} \\
MiniCPM-o 4.5 & Dense & 35.50 & 57.25 & \textbf{62.25} & \textbf{+5.00} \\
\addlinespace[3pt]
Qwen3-Omni    & MoE   & 13.25 & 34.00 & \textbf{62.50} & \textbf{+28.50} \\
Nemotron      & MoE   & 0.50  & 6.00  & \textbf{16.50} & \textbf{+10.50} \\
\bottomrule
\end{tabular}
\end{table}

\subsection{SyncRA improves general video QA on every model--benchmark combination}
\label{sec:benchmarks}

Table~\ref{tab:main-results} compares SyncRA with Vanilla SFT while holding answer data, training budget, and backbone update scope fixed.

\begin{table}[t]
\centering
\small
\setlength{\tabcolsep}{4.3pt}
\caption{\textbf{SyncRA improves general video QA on every model--benchmark combination.} Benchmark accuracy (\%) across two dense and two MoE backbones. Vanilla SFT and SyncRA: three-run mean $\pm$ sample SD; Base and Text CoT-SFT: single checkpoints. Avg-5 weights all five benchmarks equally; $\Delta$ is SyncRA minus Vanilla SFT in percentage points (unrounded means).}
\label{tab:main-results}
\begin{tabular}{@{}lrrrrrr@{}}
\toprule
Benchmark & $n$ & Base & Text CoT-SFT & Vanilla SFT & SyncRA & $\Delta$ \\
\midrule
\multicolumn{7}{@{}l}{\textit{Qwen2.5-Omni-7B (dense, full SFT)}} \\
WorldSense & 3,172 & 45.27 & 47.92 & $50.63\,\pm\,0.32$ & $\mathbf{52.74\,\pm\,0.28}$ & +2.11 \\
Daily-Omni & 1,197 & 61.74 & 63.16 & $68.84\,\pm\,0.36$ & $\mathbf{72.26\,\pm\,0.29}$ & +3.43 \\
OmniVideoBench & 1,000 & 25.10 & 35.50 & $37.77\,\pm\,0.31$ & $\mathbf{38.57\,\pm\,0.25}$ & +0.80 \\
LVOmniBench & 1,014 & 28.21 & 31.66 & $38.03\,\pm\,0.21$ & $\mathbf{39.55\,\pm\,0.30}$ & +1.51 \\
AVUT-Human & 1,733 & 64.74 & 67.86 & $67.97\,\pm\,0.31$ & $\mathbf{71.46\,\pm\,0.34}$ & +3.48 \\
\addlinespace[2pt]
Avg-5 & \textemdash & 45.01 & 49.22 & $52.65\,\pm\,0.12$ & $\mathbf{54.92\,\pm\,0.11}$ & +2.27 \\
\midrule
\multicolumn{7}{@{}l}{\textit{MiniCPM-o 4.5 (dense, full SFT)}} \\
WorldSense & 3,172 & 55.17 & 55.71 & $56.78\,\pm\,0.19$ & $\mathbf{57.61\,\pm\,0.13}$ & +0.83 \\
Daily-Omni & 1,197 & 79.62 & 79.87 & $79.67\,\pm\,0.29$ & $\mathbf{80.51\,\pm\,0.21}$ & +0.84 \\
OmniVideoBench & 1,000 & 27.80 & 33.40 & $35.53\,\pm\,0.32$ & $\mathbf{37.40\,\pm\,0.26}$ & +1.87 \\
LVOmniBench & 1,014 & 28.21 & 31.56 & $31.72\,\pm\,0.21$ & $\mathbf{31.95\,\pm\,0.26}$ & +0.23 \\
AVUT-Human & 1,733 & 78.07 & 78.59 & $78.27\,\pm\,0.23$ & $\mathbf{79.28\,\pm\,0.21}$ & +1.02 \\
\addlinespace[2pt]
Avg-5 & \textemdash & 53.77 & 55.82 & $56.39\,\pm\,0.18$ & $\mathbf{57.35\,\pm\,0.15}$ & +0.96 \\
\midrule
\multicolumn{7}{@{}l}{\textit{Qwen3-Omni-30B-A3B (MoE, 8.56\% trainable)}} \\
WorldSense & 3,172 & 55.01 & 55.39 & $56.26\,\pm\,0.21$ & $\mathbf{57.36\,\pm\,0.13}$ & +1.09 \\
Daily-Omni & 1,197 & 74.94 & 74.44 & $76.16\,\pm\,0.27$ & $\mathbf{79.09\,\pm\,0.13}$ & +2.92 \\
OmniVideoBench & 1,000 & 43.90 & 44.20 & $46.07\,\pm\,0.32$ & $\mathbf{47.07\,\pm\,0.25}$ & +1.00 \\
LVOmniBench & 1,014 & 39.94 & 39.64 & $40.89\,\pm\,0.23$ & $\mathbf{43.92\,\pm\,0.21}$ & +3.02 \\
AVUT-Human & 1,733 & 77.78 & 77.15 & $78.09\,\pm\,0.27$ & $\mathbf{79.71\,\pm\,0.23}$ & +1.62 \\
\addlinespace[2pt]
Avg-5 & \textemdash & 58.31 & 58.16 & $59.50\,\pm\,0.14$ & $\mathbf{61.43\,\pm\,0.08}$ & +1.93 \\
\midrule
\multicolumn{7}{@{}l}{\textit{Nemotron-3-Nano-Omni-30B-A3B (MoE, 11.03\% trainable)}} \\
WorldSense & 3,172 & 52.87 & 52.90 & $54.32\,\pm\,0.22$ & $\mathbf{55.65\,\pm\,0.21}$ & +1.33 \\
Daily-Omni & 1,197 & 72.60 & 70.26 & $75.80\,\pm\,0.26$ & $\mathbf{78.42\,\pm\,0.27}$ & +2.62 \\
OmniVideoBench & 1,000 & 42.20 & 43.10 & $43.47\,\pm\,0.15$ & $\mathbf{44.93\,\pm\,0.15}$ & +1.47 \\
LVOmniBench & 1,014 & 39.15 & 39.84 & $41.58\,\pm\,0.15$ & $\mathbf{42.93\,\pm\,0.15}$ & +1.35 \\
AVUT-Human & 1,733 & 72.94 & 73.11 & $73.88\,\pm\,0.30$ & $\mathbf{77.67\,\pm\,0.31}$ & +3.79 \\
\addlinespace[2pt]
Avg-5 & \textemdash & 55.95 & 55.84 & $57.81\,\pm\,0.09$ & $\mathbf{59.92\,\pm\,0.11}$ & +2.11 \\
\bottomrule
\end{tabular}
\end{table}

SyncRA achieves higher mean accuracy than Vanilla SFT in all 20 model--benchmark combinations. For each backbone, the Avg-5 gain far exceeds the run-to-run variation of either method. The gains hold under full understanding fine-tuning for the dense models and partial parameter updates for the MoE models, supporting the same supervision interface across model families and update settings. Appendix~\ref{app:full_results} provides individual-run results and integer correct counts.

Supervising reasoning text does not substitute for correspondence supervision: Text CoT-SFT trails Vanilla SFT on the Avg-5 of every backbone (on Qwen3-Omni, on all five benchmarks), while SyncRA improves over both.

\paragraph{Part of the benchmark gain is generic.}
Table~\ref{tab:attribution} compares SyncRA with the three auxiliary controls of Section~\ref{sec:supervised_relation} on all four backbones. On Avg-5, the strongest control captures 73\% of SyncRA's gain on Qwen3-Omni, 45\% on Nemotron, and 25\% on Qwen2.5-Omni, while every control falls below Vanilla SFT on MiniCPM-o 4.5. An Avg-5 gain alone therefore cannot be attributed to correct pairing. On AllFour, the strongest control is always Fixed time codes, which supplies a shared time target and captures 35--40\% of the gain. SyncRA exceeds the strongest control on both metrics for every backbone, and only the AllFour interval on MiniCPM-o 4.5 contains zero. We therefore present the 20 improved combinations as a general QA gain with a generic component, and the pairing-specific claims rest on the swap diagnostic (Appendix~\ref{app:attribution}). The AllFour advantage holds after conditioning on original-version accuracy, in open-ended answers, and under larger answer-only budgets (Appendix~\ref{app:robustness}).

\begin{table}[t]
\centering
\small
\setlength{\tabcolsep}{3.6pt}
\caption{\textbf{Gain attribution against the auxiliary controls.} Avg-5 is the three-run mean; AllFour uses the diagnostic checkpoints (\%). $\Delta$ is SyncRA minus the strongest control (underlined), with Welch 95\% intervals over three runs (Avg-5) or paired source-video bootstrap intervals (AllFour).}
\label{tab:attribution}
\begin{tabular}{@{}llrrrrrrl@{}}
\toprule
& & Vanilla & Permuted & Fixed time & Clip-level & & & \\
Metric & Backbone & SFT & pairs & codes & AV & SyncRA & $\Delta$ & 95\% interval \\
\midrule
Avg-5 & Qwen2.5-Omni  & 52.65 & \underline{53.22} & 52.47 & 53.18 & \textbf{54.92} & $+1.70$ & $[1.38,2.02]$ \\
 & MiniCPM-o 4.5 & 56.39 & 55.21 & \underline{55.32} & 54.60 & \textbf{57.35} & $+2.03$ & $[1.65,2.41]$ \\
 & Qwen3-Omni    & 59.50 & 60.85 & 60.39 & \underline{60.91} & \textbf{61.43} & $+0.52$ & $[0.36,0.68]$ \\
 & Nemotron      & 57.81 & 58.55 & \underline{58.77} & 58.72 & \textbf{59.92} & $+1.15$ & $[0.92,1.38]$ \\
\midrule
AllFour & Qwen2.5-Omni  & 11.00 & 10.25 & \underline{14.50} & 11.25 & \textbf{20.75} & $+6.25$ & $[2.25,10.25]$ \\
 & MiniCPM-o 4.5 & 57.25 & 55.50 & \underline{59.00} & 56.25 & \textbf{62.25} & $+3.25$ & $[-0.50,+7.00]$ \\
 & Qwen3-Omni    & 34.00 & 31.00 & \underline{44.75} & 31.75 & \textbf{62.50} & $+17.75$ & $[13.00,22.75]$ \\
 & Nemotron      & 6.00  & 6.75  & \underline{10.25} & 7.50  & \textbf{16.50} & $+6.25$ & $[2.50,10.25]$ \\
\bottomrule
\end{tabular}
\end{table}

\section{Ablation Study and Analysis}
\label{sec:ablation}

We next examine why the supervision works and how it changes the backbone states. Alignment-loss, prior-objective, supervision-depth, and quantitative representation analyses use Qwen3-Omni Run~1; cross-backbone objective comparisons report three-run benchmark means, and qualitative heatmaps cover all four backbones.

\subsection{Local contrastive supervision drives the gains}
\label{sec:supervised_relation}
\label{sec:loss_ablation}

\paragraph{The pairing relation drives the AllFour gain.}
Aggregate QA accuracy alone conceals the benefit of supervising the correct local relation. We compare SyncRA with incorrect within-video pairs (Permuted pairs), shared timestamp targets (Fixed time codes), and global audio--video association (Clip-level AV). Their Daily-Omni accuracies are close to SyncRA's, yet their AllFour scores diverge (Table~\ref{tab:objective-ablation}). Even the strongest alternative temporal target, Fixed time codes, trails direct local pairing by 17.75 percentage points on AllFour. The supervised relation itself, not generic temporal or cross-modal alignment, drives the AllFour gain. SyncRA also achieves the highest Avg-5 among these objectives on all four backbones (Appendix~\ref{app:full_results}).

\paragraph{Competing intervals are necessary.}
The alignment objective must distinguish the matching moment from other times in the same recording. We therefore compare positive-only cosine alignment, independent pairwise logistic ranking, and InfoNCE's jointly normalized interval classification, keeping the same co-occurring positive pairs and training setup. InfoNCE achieves the highest AllFour accuracy, reaching 62.50\% versus 46.25\% for pairwise ranking, despite similar Daily-Omni scores (Table~\ref{tab:objective-ablation}). The gain therefore comes from both the local pairing target and its contrastive formulation. Objective definitions, the loss rationale, and paired confidence intervals appear in Appendix~\ref{app:loss_ablation}.

\paragraph{Comparison with prior synchrony objectives.}
We port the auxiliary losses of \citet{korbar2018cooperative}, CAV-MAE Sync \citep{araujo2025cavmaesync}, and AVAF-Net \citep{zhao2025avafnet} to Qwen3-Omni. Only the auxiliary loss changes, and all three losses converge (Appendix~\ref{app:ported}). Korbar et al.\ and AVAF-Net are statistically indistinguishable from Vanilla SFT on AllFour. CAV-MAE Sync reaches 52.50\%, 10.00 points below SyncRA with a paired 95\% interval of $[5.75,14.25]$ (Table~\ref{tab:objective-ablation}). All four objectives act on the same block, so the form of the objective decides whether pairing-following improves. The ordering is consistent with three factors: interval granularity (Clip-level AV), the InfoNCE form (cosine-only, pairwise ranking), and same-video negatives (CAV-MAE Sync).

\paragraph{AllFour rises with alignment strength.}
Table~\ref{tab:objective-ablation} also reports same-interval retrieval at block 36 for all twelve configurations, including two other auxiliary weights (Appendix~\ref{app:aux_weight}). The five configurations whose R@1 stays at the Vanilla level (11.8--13.9\%) have AllFour between 31.00\% and 34.50\%. Above this level, AllFour increases with R@1, up to 62.50\% at 96.9\% R@1.

\subsection{SyncRA aligns the backbone's own states in time}
\label{sec:representation}
\label{sec:layer_analysis}

SyncRA strengthens temporal correspondence in the backbone's own representations, and the change lasts to the final layer. At the final block of Qwen3-Omni, 12 blocks after the supervised block, unprojected same-interval retrieval reaches 31.06\% R@1 for SyncRA, against 3.44\% for Vanilla SFT and at most 7.49\% for the three auxiliary controls (Appendix~\ref{app:depth_analysis}). The aligned states track the current input position rather than semantic matching: after a two-interval audio delay, SyncRA retrieves the current interval at 96.45\% R@1 at the supervised block but its content origin at only 0.01\% (Appendix~\ref{app:audio_controls}). Retrieval follows when, not what.

\begin{figure}[t]
\centering

\begin{minipage}[t]{0.59\linewidth}
\vspace{0pt}
\centering
\small
\renewcommand{\arraystretch}{1.02}
\setlength{\tabcolsep}{3pt}
\captionof{table}{%
\textbf{Local contrastive supervision matters.}
Qwen3-Omni, Run~1; scores in \%. R@1: unprojected same-interval
retrieval at block 36 (targets and ported objectives: refreshed
pipeline, differs from Table~\ref{tab:representation} by $\le$0.1).
}
\label{tab:objective-ablation}
\label{tab:loss-ablation-short}
\begin{tabular}{@{}lrrr@{}}
\toprule
Training objective & AllFour $\uparrow$ & R@1 $\uparrow$ & Daily-Omni $\uparrow$ \\
\midrule
Vanilla SFT & 34.00 & 13.4 & 75.86 \\
\midrule
\multicolumn{4}{@{}l}{\textit{Alternative supervision targets}} \\
Permuted pairs   & 31.00 & 12.7 & 78.78 \\
Fixed time codes & 44.75 & 22.6 & 78.45 \\
Clip-level AV    & 31.75 & 13.9 & 78.70 \\
\multicolumn{4}{@{}l}{\textit{Alternative alignment losses}} \\
Cosine-only      & 40.50 & 17.9 & 78.20 \\
Pairwise ranking & 46.25 & 27.5 & 78.36 \\
\multicolumn{4}{@{}l}{\textit{Prior synchrony objectives (ported)}} \\
Korbar et al.    & 33.25 & 12.1 & 77.27 \\
CAV-MAE Sync     & 52.50 & 53.6 & 78.36 \\
AVAF-Net         & 34.50 & 11.8 & 77.94 \\
\multicolumn{4}{@{}l}{\textit{Other auxiliary weights}} \\
SyncRA, $\lambda=0.03$ & 58.00 & 90.9 & 78.53 \\
SyncRA, $\lambda=0.3$  & 58.75 & 93.7 & 78.36 \\
\midrule
\textbf{SyncRA (local InfoNCE)} & \textbf{62.50} & \textbf{96.9} & 78.95 \\
\bottomrule
\end{tabular}

\end{minipage}\hfill
\begin{minipage}[t]{0.3771\linewidth}
\vspace{0pt}
\centering
\includegraphics[width=\linewidth]{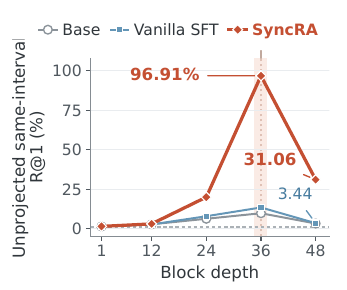}
\captionof{figure}{%
\textbf{Correspondence across depth in \mbox{Qwen3-Omni}.}
}
\label{fig:depth-profile}
\vspace{-5pt}
\small
\setlength{\tabcolsep}{3pt}
\captionof{table}{%
\textbf{Activation patching} on Qwen3-Omni (normalized effect).
}
\label{tab:patching-main}
\begin{tabular}{@{}lrr@{}}
\toprule
States copied & Vanilla & SyncRA \\
\midrule
Video, both windows   & 0.99 & 0.99 \\
Video, cue window     & 0.91 & 0.97 \\
Video, partner window & 0.12 & 0.02 \\
Audio, cue window     & 0.32 & 0.46 \\
\bottomrule
\end{tabular}
\end{minipage}

\end{figure}

\paragraph{The answer reads the cue-time states.}
In activation patching, we run Qwen3-Omni on a swap-test clip and on its video-exchanged version, copy a chosen subset of hidden states from the second run into the first, and measure how far the answer moves toward that of the exchanged input, where 1 equals running the exchanged input itself (Table~\ref{tab:patching-main}; Appendix~\ref{app:patching}). Copying only the video states of the cue window, in which the cue is spoken, flips 93\% of SyncRA's answers and 83\% of Vanilla SFT's. The video states of the partner window, exchanged with it, have an effect of 0.02 for SyncRA and 0.12 for Vanilla SFT, and the audio states of the cue window, whose audio input is identical in the two runs, 0.46 and 0.32. Both models read the cue-time states; SyncRA reads them more selectively, and its audio states at that time carry more of the simultaneous visual information.

\begin{figure}[t]
\centering
\makebox[0.48\linewidth][c]{\small Qwen3-Omni (MoE), block 36}\hfill
\makebox[0.48\linewidth][c]{\small Qwen2.5-Omni (dense), block 21}\par
\smallskip
\makebox[0.24\linewidth][c]{\small Vanilla SFT}%
\makebox[0.24\linewidth][c]{\small SyncRA}%
\makebox[0.24\linewidth][c]{\small Vanilla SFT}%
\makebox[0.24\linewidth][c]{\small SyncRA}\par
\includegraphics[width=1\linewidth]{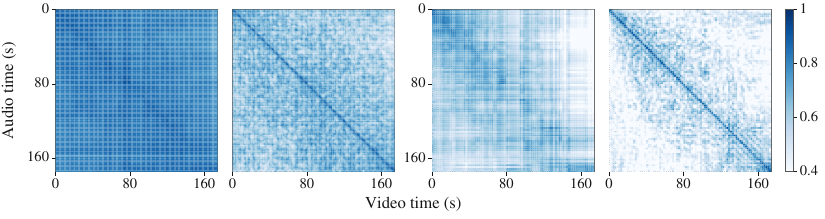}
\caption{
\textbf{Temporal correspondence in dense and MoE backbone states.}
Cosine similarity between unprojected interval-mean audio (rows) and
video (columns) states on a fixed validation clip (Appendix~\ref{app:cross-backbone-heatmaps}); all panels share a 0.4--1 color scale.
}
\label{fig:main-similarity}
\end{figure}

\paragraph{Correspondence at the supervised block.}
Figure~\ref{fig:main-similarity} compares native interval-pooled states on the same fixed clip: both Qwen3-Omni and Qwen2.5-Omni develop a more concentrated same-time diagonal after SyncRA training. Aggregate retrieval quantifies this change. The backbone already knows the time: elapsed time is linearly decodable from the frozen released checkpoint, at 8.996~s (audio) and 5.675~s (video) MAE against 30.816~s and 31.041~s under shuffled labels, and answer-only fine-tuning only sharpens this readout (Appendix~\ref{app:linear_time}). What it lacks is cross-modal alignment---direct same-interval matching sits at 13.39\% unprojected R@1 on 100 Qwen3-Omni validation videos. SyncRA supplies exactly this missing piece, raising R@1 to 96.91\%. The gain peaks at the supervised block and remains visible in the final block (Figure~\ref{fig:depth-profile}).

\paragraph{Selecting the supervision block.}
The supervision block is chosen per backbone by probing frozen Base states before QA fine-tuning (Section~\ref{sec:method}; Appendix~\ref{app:layer_selection}), and the selected relative depths differ across backbones. Among the two depths we tested, training at Qwen3-Omni block 24 or 36 both improves AllFour over Vanilla SFT, while the supervision location shapes where in the network correspondence concentrates (Table~\ref{tab:supervision-depth} in Appendix~\ref{app:depth_analysis}).

\section{Related Work}
\label{sec:related}

\paragraph{Audio--visual correspondence and synchronization.} A long line of work learns multisensory representations from audio--visual synchrony, using same-time frame--audio pairs as positives and pairs from other videos or shifted times as negatives \citep{arandjelovic2017look,korbar2018cooperative,owens2018multisensory}; CAV-MAE Sync explicitly refines frame--audio alignment \citep{araujo2025cavmaesync}, and AVAF-Net applies a temporal-alignment contrastive loss to audio--visual question answering \citep{zhao2025avafnet}. SyncRA instead contrasts same-input-time pairs inside an answering model, and its retrieval analysis measures this temporal organization in the model's own states rather than in independently encoded content.

\paragraph{Temporal supervision in omni-modal reasoning.} Daily-Omni and AVUT emphasize evaluation that requires both audio and vision \citep{zhou2025dailyomni,yang2025avut}. LatentOmni combines temporal InfoNCE, sensory feature supervision, and text prediction in interleaved text--latent reasoning, and it aligns extra latent states that it generates at inference \citep{dai2026latentomni}; SyncRA instead aligns the answering model's own intermediate states during QA fine-tuning, leaves inference unchanged, and tests whether answers track changes in the current pairing. TimeLens studies timestamp representations for continuous video temporal grounding \citep{zhang2026timelens}; our diagnostic asks for scene selection during a spoken cue.

\paragraph{Intermediate representation supervision.} REPA and iREPA align diffusion-model states to pretrained visual features \citep{yu2025repa,singh2026irepa}; SyncRA instead obtains supervision from co-occurring input intervals and jointly trains the answering model's audio and visual states.

\section{Conclusion}
\label{sec:conclusion}

Omni-modal models often fail to connect what they hear with what they see at the same moment. SyncRA makes this local audio--visual correspondence an explicit target during QA post-training, improving correspondence-sensitive answering across dense and MoE backbones and raising mean accuracy on every evaluated model--benchmark combination. Its scope is input-time correspondence between the two streams; narration referents, off-screen sound semantics, and desynchronization detection are outside it. Controlled comparisons show that part of the benchmark gain is generic to auxiliary training and that similar aggregate QA scores can conceal large differences in whether answers follow the current pairing. Local temporal correspondence is a distinct, directly supervisable capability, and targeted representation supervision offers a general way to strengthen it.
\label{maintextend}

\clearpage
\section*{AI Use Statement}
We used LLMs to improve manuscript grammar and clarity and assist routine code modifications. The authors reviewed and verified all AI-assisted revisions, corrected them as needed, and take responsibility for the manuscript, code, and results.

\section*{Reproducibility Statement}
Appendices detail data splits, optimization, checkpoint and layer selection, backbone configurations, and evaluation. They document diagnostic construction, human verification, scoring, confidence intervals, representation analyses, and per-run benchmark counts.
\bibliography{references}
\bibliographystyle{iclr2027_conference}
\clearpage
\appendix


\section{Limitations}
\label{sec:limitations}

SyncRA targets local input-time correspondence between the audio and visual streams; narration referents, off-screen sound semantics, and desynchronization detection are outside this scope (Appendices~\ref{app:sync_judgment} and~\ref{app:interval_audit}). Within this scope, our study has four limitations. First, the swap test covers 100 source videos. Its open-ended version preserves SyncRA's lead over the strongest control under two LLM judges, with lower scores for every model (Appendix~\ref{app:robustness}). Second, the quantitative mechanism analyses in Section~\ref{sec:ablation} use Qwen3-Omni Run~1 checkpoints, in several cases a single training run; for the other three backbones we report AllFour gains and heatmaps, but no patching. Patching uses the instances that each model answers correctly, so the two models are compared on different instance sets. We have not removed the alignment from a trained SyncRA model to test whether the AllFour gain disappears, so the link between alignment strength and AllFour rests on the comparison across configurations. Third, the smallest gain occurs on MiniCPM-o 4.5 (+5.00 points), whose Base model already tracks pairing best (AllFour 35.50\%); how much SyncRA adds on better-aligned models remains open. Fourth, the main experiments use the same compact 10K-question corpus. On Qwen3-Omni, the AllFour advantage holds with 3$\times$ data and with up to 8 epochs (Appendix~\ref{app:robustness}); we have not tested beyond 30K questions.

\section{Training and Evaluation Details}
\label{app:training}

\subsection{Data and answer supervision}
All four backbones use the fixed, balanced 10K-example subset of OmniVideo-100K described in Section~\ref{sec:setup}, containing 10,000 audio--video question--answer examples from 1,782 source videos. Clips last 60--180 seconds. The subset contains 5,000 multiple-choice and 5,000 open-ended questions. Examples were selected from available source files that passed media-quality checks, balancing task and answer format with selection seed 1. Table~\ref{tab:app_training_data} gives the source-disjoint split and task composition. Source identity is the released YouTube video ID.

\begin{table}[htbp]
\centering\small
\caption{Training data. Each of the ten task categories contributes 1,000 questions to the complete 10,000-question subset. Source videos are disjoint between training and validation.}
\label{tab:app_training_data}
\begin{tabular}{lrr}
\toprule
Split & Questions & Source videos\\
\midrule
Training & 9,500 & 1,367\\
Validation & 500 & 415\\
Total & 10,000 & 1,782\\
\bottomrule
\end{tabular}\qquad
\begin{tabular}{l}
\toprule
Task categories\\
\midrule
Causal reasoning; comparison\\
Context understanding; event sequence ordering\\
Fine-grained perception; future prediction\\
Hypothetical reasoning; scene transformation detection\\
Sentiment analysis; summarization\\
\bottomrule
\end{tabular}
\end{table}

Vanilla SFT, SyncRA, Permuted pairs, Fixed time codes, and Clip-level AV use the same media, questions, options, final answers, and a fixed example order. A backbone's native chat template is held fixed across its training objectives. Supervision covers final answer tokens; supplied reasoning traces and evidence-chain markers are removed from this training view. SyncRA obtains its interval correspondences from native audio--video timing information. It adds no event-boundary labels, external representation teacher, or inference branch.

\subsection{Optimization and trainable parameters}
\label{app:optimization}
Vanilla SFT, SyncRA, Permuted pairs, Fixed time codes, and Clip-level AV each use three independent training runs with different random seeds and the common settings in Table~\ref{tab:app_optimizer}. Each of these runs selects its checkpoint by the lowest validation mean answer-token negative log-likelihood (NLL), evaluated at the end of each epoch; ties select the earlier epoch. Text CoT-SFT selection is specified in Appendix~\ref{app:text_cot}. Benchmark scores and post-training retrieval diagnostics do not enter checkpoint selection. The auxiliary projection is removed after training, and evaluation uses the model's native answer-generation path. All experiments use PyTorch 2.5.1+cu124; each training run uses two NVIDIA H200 GPUs with 141 GB of memory per GPU.

\begin{table}[htbp]
\centering\small
\caption{Optimization settings for Vanilla SFT, SyncRA, and the three auxiliary controls. Runs 1--3 use different random seeds. Qwen3-Omni diagnostic checkpoints for these methods are from Run~1; the fixed checkpoints for the other backbones are described in Appendix~\ref{app:behavior_metrics}.}
\label{tab:app_optimizer}
\begin{tabularx}{\linewidth}{@{}lX@{}}
\toprule
Setting & Value\\
\midrule
Optimizer & AdamW; learning rate $10^{-5}$; $(\beta_1,\beta_2)=(0.9,0.95)$; $\epsilon=10^{-8}$; weight decay 0.1\\
Schedule & Linear decay; 80 warmup steps\\
Budget & 2 epochs; 792 optimizer steps per epoch; 1,584 steps total\\
Effective optimization batch & 12 examples: 2 processes $\times$ microbatch 1 $\times$ accumulation 6\\
Precision and stabilization & bfloat16; float32 normalized contrastive features, similarities, and contrastive cross-entropy; gradient clipping 1.0; gradient checkpointing\\
Data order & Fixed example order; no additional shuffling\\
Validation and selection & End of each epoch; minimum validation answer-token NLL; earlier epoch breaks ties\\
Auxiliary objectives & Shared projection dimension 64; temperature 0.07; loss weight 0.1\\
\bottomrule
\end{tabularx}
\end{table}

\begin{table}[htbp]
\centering\small
\caption{Trainable backbone parameters. Parameter counts exclude the auxiliary projection. Each backbone uses the same update scope across the compared objectives.}
\label{tab:app_trainable}
\begin{tabularx}{\linewidth}{@{}p{0.24\linewidth}X@{}}
\toprule
Backbone & Update scope\\
\midrule
Qwen2.5-Omni-7B & Full understanding/Thinker fine-tuning, including text, visual, and audio modules; speech generation is not trained.\\
MiniCPM-o 4.5 & Full understanding fine-tuning; text-to-speech is disabled.\\
Qwen3-Omni-30B-A3B & Routed experts and router gates are frozen; other permitted modules are trained. Trainable backbone parameters: 2,715,593,328 / 31,719,205,488 (8.56\%).\\
Nemotron-3-Nano-Omni-30B-A3B & Routed experts are frozen; routers, shared experts, and other permitted modules are trained. Trainable backbone parameters: 3,640,738,752 / 33,015,546,816 (11.03\%).\\
\bottomrule
\end{tabularx}
\end{table}

\begin{table}[htbp]
\centering\small
\caption{Pretrained model identifiers and exact revisions used for initialization.}
\label{tab:app_revisions}
\begin{tabularx}{\linewidth}{@{}lX@{}}
\toprule
Backbone & Identifier and revision\\
\midrule
Qwen2.5-Omni & \texttt{Qwen/Qwen2.5-Omni-7B}\newline\texttt{ae9e1690543ffd5c0221dc27f79834d0294cba00}\\
MiniCPM-o 4.5 & \texttt{openbmb/MiniCPM-o-4\_5}\newline\texttt{503e754207c94da6bb26850b4469f367c9ea3582}\\
Qwen3-Omni & \texttt{Qwen/Qwen3-Omni-30B-A3B-Instruct}\newline\texttt{26291f793822fb6be9555850f06dfe95f2d7e695}\\
Nemotron & \texttt{nvidia/Nemotron-3-Nano-Omni-30B-A3B-}\newline\texttt{Reasoning-BF16}\newline\texttt{e5e9932441de940c9a62185c870ea5bcd4cd24e2}\\
\bottomrule
\end{tabularx}
\end{table}

\subsection{Native temporal layouts and media processing}
\label{app:media}
Temporal intervals are the model processor's sampling units, rather than annotated semantic events. Audio and video tokens are assigned to the same elapsed-time grid using native timestamps, interval boundaries, or temporal patch metadata. For each interval, token states are mean pooled separately by modality. The valid set contains intervals with nonempty audio and visual token sets. This interface also accommodates models that arrange the modality tokens in separate blocks: interval membership follows time metadata rather than an assumption about alternating sequence positions.

\begin{table}[htbp]
\centering\small
\caption{Backbone adaptation and training media. All backbones receive full 16 kHz audio. Layer indices are one-indexed and refer to block outputs. $d$ is the hidden dimension; the projection has $64d$ trainable weights and no bias.}
\label{tab:app_adaptation}
\begin{tabularx}{\linewidth}{@{}lrrrX@{}}
\toprule
Backbone & $d$ & Layer/depth & $|W|$ & Training video and interval grid\\
\midrule
Qwen2.5-Omni & 3,584 & 21/28 & 229,376 & 2 FPS, at most 256 frames; native 2 s intervals\\
MiniCPM-o 4.5 & 4,096 & 9/36 & 262,144 & 1 FPS, at most 180 one-second chunks; native 1 s intervals\\
Qwen3-Omni & 2,048 & 36/48 & 131,072 & 2 FPS, at most 256 frames; temporal patch size divided by actual sampling FPS\\
Nemotron & 2,688 & 26/52 & 172,032 & 2 FPS, at most 256 frames; timestamp-defined 2 s intervals\\
\bottomrule
\end{tabularx}
\end{table}

For both Qwen models, the configured per-frame minimum and maximum pixel budgets are 100,352 and 602,112, and the total pixel budget is 19,267,584. Actual processor outputs determine the temporal grid. Qwen3 intervals are approximately one second at 2 FPS; frame-budget resampling can change their duration on longer inputs. Nemotron groups video tubelets and audio tokens by timestamps; its selected 26th block is a Mamba2 block. Training media and benchmark long-video inputs follow their respective processing protocols.

\subsection{Layer selection on frozen Base models}
\label{app:layer_selection}
We select one layer per backbone before fine-tuning, using its released Base weights in evaluation mode. Candidate layers are fixed at approximately one-quarter, one-half, and three-quarters of the model depth. For each source video, the first question in the fixed example order defines a prompt-only audio--video forward pass, without an appended answer or reasoning trace. We cache the mean-pooled modality states at every candidate block output. A usable video must pass layout validation and contain at least two valid intervals.

\begin{table}[htbp]
\centering\small
\caption{Frozen-Base layer-selection data and selected layers. Counts are usable independent source videos for probe fitting and validation, respectively. Training itself retains the 9,500/500-question split.}
\label{tab:app_layer_selection}
\begin{tabularx}{\linewidth}{@{}lccXr@{}}
\toprule
Backbone & Candidates & Fit/validate & Probe media scope & Selected\\
\midrule
Qwen2.5-Omni & 7, 14, 21 & 200/415 & Full training clip; 2 FPS, at most 256 frames; native 2 s grid & 21\\
MiniCPM-o 4.5 & 9, 18, 27 & 200/415 & First 60 s; native 1 s audio--video intervals & 9\\
Qwen3-Omni & 12, 24, 36 & 198/408 & Full training clip; 2 FPS, at most 256 frames; actual temporal grid & 36\\
Nemotron & 13, 26, 39 & 200/415 & First 120 s; 2 FPS; native 2 s intervals & 26\\
\bottomrule
\end{tabularx}
\end{table}

The fitting set consists of the first 200 independent training-source videos in this fixed order, and validation uses all 415 validation-source videos. Qwen3's extraction excludes two training and seven validation videos that fail the audio--video interleaving-layout check, leaving 198/408 usable videos. These counts concern this probe extraction; the SFT split is unchanged. MiniCPM and Nemotron use the fixed probe-specific clip scopes in Table~\ref{tab:app_layer_selection}.

For each candidate $\ell$, we fit a separate bias-free shared projection $P_\ell\in\mathbb R^{64\times d}$ on the frozen states. Candidate projections use the same initialization seed, 1. Only $P_\ell$ is updated: Adam, learning rate $10^{-3}$, temperature 0.07, one video per update, fixed video order, and 50 epochs. The objective is the same within-video symmetric interval InfoNCE used by SyncRA, with L2-normalized projected states. There is no answer loss in probe fitting. Validation uses the probe at the end of epoch 50.

Selection scores pool bidirectional hits over all validation queries:
\begin{equation}
R_\ell=\frac{\sum_{i=1}^{N}(H^{a\to v}_{i,\ell}+H^{v\to a}_{i,\ell})}{2\sum_{i=1}^{N}K_i},
\label{app:eq_selection}
\end{equation}
where $K_i$ is the valid interval count and $H$ is the number of same-interval top-1 hits in the indicated direction. A candidate is eligible when $R_\ell\in[c+0.05,0.90)$: the lower bound requires signal five percentage points above the chance reference $c$, and the upper bound reserves learning headroom. Among eligible candidates, the largest $R_\ell$ is selected. The rule makes no selection when no candidate is eligible. Each of the four backbones has an eligible candidate.

The chance reference $c$ is computed per backbone from its interval counts: 1.92\% for Qwen2.5-Omni, 1.67\% for MiniCPM-o 4.5, 1.02\% for Qwen3-Omni, and 2.25\% for Nemotron. All selection scores $R_\ell$ use the query-pooled definition in Eq.~\eqref{app:eq_selection}.

For the main experiments, the selected layer is fixed across
auxiliary objectives and training runs. Fine-tuning starts from Base with a newly initialized $W$; the fitted layer-selection projections $P_\ell$ are not reused. Qwen3-Omni projected-state diagnostics use the SyncRA $W$ from the selected Run~1 checkpoint.

\begin{table}[htbp]
\centering\small
\setlength{\tabcolsep}{5pt}
\caption{Frozen-Base shared-probe validation R@1 (\%) for all twelve candidate layers, using Eq.~\eqref{app:eq_selection}. All candidates satisfy the eligibility band; bold marks the selected block and score for each backbone.}
\label{tab:candidates}
\begin{tabular}{@{}lrrrrrr@{}}
\toprule
& \multicolumn{2}{c}{Quarter depth} & \multicolumn{2}{c}{Half depth} & \multicolumn{2}{c}{Three-quarter depth}\\
\cmidrule(lr){2-3}\cmidrule(lr){4-5}\cmidrule(lr){6-7}
Backbone & Block & R@1 & Block & R@1 & Block & R@1\\
\midrule
Qwen2.5-Omni & 7 & 11.44 & 14 & 16.47 & \textbf{21} & \textbf{20.00}\\
MiniCPM-o 4.5 & \textbf{9} & \textbf{78.61} & 18 & 74.39 & 27 & 36.44\\
Qwen3-Omni & 12 & 13.07 & 24 & 28.20 & \textbf{36} & \textbf{28.30}\\
Nemotron & 13 & 13.30 & \textbf{26} & \textbf{14.06} & 39 & 12.91\\
\bottomrule
\end{tabular}
\end{table}

Table~\ref{tab:candidates} shows that the strongest probe correspondence occurs at different relative depths across backbones: three-quarters for both Qwen models, one-quarter for MiniCPM-o, and one-half for Nemotron. This selection procedure adapts the supervision location to each architecture while retaining the same temporal alignment objective.

\subsection{Auxiliary loss implementation}
\label{app:losses}
\label{app:objectives}
For a video with $K$ valid intervals, let $a_k,v_k\in\mathbb R^d$ be the pooled states. The shared projection $W\in\mathbb R^{64\times d}$ has no bias. Projection uses the weight dtype, then casts its output to float32 before L2 normalization with $\epsilon=10^{-8}$. Similarity logits and contrastive cross-entropy are float32. Define
\begin{equation}
z^a_k=\frac{Wa_k}{\max(\|Wa_k\|_2,\epsilon)},\qquad
z^v_k=\frac{Wv_k}{\max(\|Wv_k\|_2,\epsilon)},\qquad
s_{kj}=\frac{(z^a_k)^\top z^v_j}{\tau},\quad \tau=0.07.
\label{app:eq_projection}
\end{equation}
For an assigned bijective pairing $\pi$, the implemented loss is
\begin{equation}
\mathcal L_{\mathrm{pair}}(\pi)=-\frac{1}{2K}\left[
\sum_{k=1}^{K}\log\frac{e^{s_{k,\pi(k)}}}{\sum_j e^{s_{kj}}}
+\sum_{j=1}^{K}\log\frac{e^{s_{\pi^{-1}(j),j}}}{\sum_k e^{s_{kj}}}\right].
\label{app:eq_pairing}
\end{equation}
Each direction averages its $K$ query losses, and the two directional means are averaged. SyncRA uses $\pi(k)=k$. The sample's training objective is $\mathcal L_{\mathrm{answer}}+0.1\mathcal L_{\mathrm{SyncRA}}$; its auxiliary loss is computed before the common SFT gradient-accumulation procedure. When $K<2$, the sample uses answer supervision alone. Auxiliary gradients update $W$ and the permitted backbone parameters, while answer-loss gradients continue through the subsequent layers.

\paragraph{Permuted pairs.}
A random generator is seeded deterministically from the source-video ID. It shuffles interval indices until obtaining a derangement, so $\pi(k)\ne k$. The implementation permits at most 10,000 attempts and raises an error if none succeeds. Audio-to-video targets are $\pi$, and video-to-audio targets are $\pi^{-1}$. Media, answers, candidate intervals, projection size, and training settings match SyncRA. This control changes the designated positive relationship while preserving the same within-video candidate set.

\paragraph{Fixed time codes.}
Let $t_k$ be the starting time in the input clip of valid interval $k$. In the native-grid implementation this is the original valid grid index multiplied by seconds per interval, rather than the index after filtering invalid intervals. For 32 log-spaced periods $p_j$ from 4 to 512 seconds, define the fixed unit-norm code
\begin{equation}
c(t)=\frac{1}{\sqrt{32}}\left[
\sin(2\pi t/p_1),\ldots,\sin(2\pi t/p_{32}),
\cos(2\pi t/p_1),\ldots,\cos(2\pi t/p_{32})\right].
\label{app:eq_clock_code}
\end{equation}
Writing $C_{k,:}=c(t_k)$ and stacking normalized projected modality features into $Z^a,Z^v$, the loss is
\begin{equation}
\mathcal L_{\mathrm{clock}}=\tfrac12\left[
\mathrm{CE}(Z^aC^\top/\tau,y)+\mathrm{CE}(Z^vC^\top/\tau,y)\right],\quad y_k=k.
\label{app:eq_clock}
\end{equation}
Both modalities use the same projection $W$ and the same fixed codes $C$. Their losses separately discriminate the codes of the current video's valid intervals, without directly comparing audio and visual feature vectors. This supplies a common temporal target; the contrast with SyncRA is between fixed timestamp-code supervision and learned pairwise correspondence.

\paragraph{Clip-level AV.}
For each current video, average its valid interval states separately by modality, then project and normalize: $\bar z^a=\mathrm{norm}(W K^{-1}\sum_k a_k)$ and likewise for video. The interval average gives each interval equal weight. Differentiable all-gather collects one clip per training process. Let $B=2$ be this current candidate count, $g_i$ its source IDs, $u_{ij}=(\bar z_i^a)^\top\bar z_j^v/\tau$, and $\mathcal P_i=\{j:g_j=g_i\}$. The loss is
\begin{equation}
\begin{split}
\mathcal L_{\mathrm{global}}=\frac{1}{2B}\sum_{i=1}^{B}\bigg[
&\log\sum_{j=1}^{B}e^{u_{ij}}-\log\sum_{j\in\mathcal P_i}e^{u_{ij}}\\
+&\log\sum_{j=1}^{B}e^{u_{ji}}-\log\sum_{j\in\mathcal P_i}e^{u_{ji}}\bigg].
\end{split}
\label{app:eq_global}
\end{equation}
Repeated source IDs all enter the positive set. Each process evaluates the same gathered loss; differentiable gather and distributed gradient averaging implement the common update. The effective optimization batch is 12, but accumulation does not enlarge the contrastive candidate pool beyond the two current videos. Fixed time codes and Clip-level AV also enter the answer-training objective with weight 0.1.

\subsection{Text CoT-SFT baseline}
\label{app:text_cot}
Text CoT-SFT starts from the released backbone weights (Table~\ref{tab:app_revisions}) and shares Vanilla SFT's data split, optimization settings, two-epoch budget (1,584 steps), and trainable modules (Tables~\ref{tab:app_optimizer} and~\ref{tab:app_trainable}). Targets are the supplied text reasoning followed by a newline and \mbox{\texttt{Final answer: \{answer\}}}. Validation target-token NLL covers the complete target on all 500 validation examples; all four backbones select epoch 2. Training retains complete prompts and targets without a fixed text-token cap or truncation, using the media limits in Table~\ref{tab:app_adaptation} (at most 180\,s).

\begin{table}[H]
\centering\small
\caption{Text CoT-SFT benchmark evaluation prompts (no examples). Mode and template settings apply during both training and evaluation.}
\label{tab:app_cot_prompts}
\begin{tabularx}{\linewidth}{@{}l>{\raggedright\arraybackslash}X>{\raggedright\arraybackslash}X@{}}
\toprule
Backbone & Evaluation instruction & Mode/template setting\\
\midrule
Qwen2.5-Omni & Direct option-letter answer & No additional thinking mode\\
MiniCPM-o 4.5 & Step-by-step reasoning;\newline final line \mbox{\texttt{Final answer: X}} & \texttt{enable\_thinking=False}; \texttt{use\_tts\_template=False}\\
Qwen3-Omni & Step-by-step reasoning;\newline final line \mbox{\texttt{Final answer: X}} & Instruct weights\\
Nemotron & Neutral question and options & \texttt{enable\_thinking=False}\\
\bottomrule
\end{tabularx}
\end{table}

Qwen2.5 training uses neutral question-and-options text. Base, Vanilla SFT, and SyncRA retain their direct-answer evaluation prompts.

\subsection{Benchmark inputs, generation, and statistical summaries}
\label{app:benchmark_protocol}
Benchmark evaluations use fixed questions and options, native model templates, greedy generation, EOS stopping, and a maximum of 1,280 new tokens. This budget applies only to evaluation generation. Each checkpoint generates one response per question. Accuracy is $100C/n$, where $C$ is the integer correct count and $n$ is the fixed effective denominator. Generation failures, parsing failures, and incorrect answers remain in the denominator.

Base, Vanilla SFT, SyncRA, and the three auxiliary controls use a case-insensitive regular-expression parser. It accepts a complete direct option answer or the last valid \texttt{Final answer:} field, with allowed punctuation followed by whitespace or the response boundary. Bare letters in reasoning are ignored; neither semantic matching nor a judge model is used. Non-choice time outputs use a numeric/time parser.

For Text CoT-SFT, Qwen2.5 accepts a complete direct answer or the last explicit \texttt{Final answer:} field anywhere in the response. The other backbones accept a complete single-line answer or a last-line field labeled \texttt{Final answer:}, \texttt{Answer:}, or \texttt{Correct answer:} (including an inline \texttt{Final answer:} suffix). Letters and option text must agree; without a letter, the field must uniquely match a complete option, allowing case, whitespace, and limited punctuation differences. Ambiguous or conflicting answers are incorrect. Outputs reaching the generation cap are parsed from the saved text under the same rules.

\begin{table}[htbp]
\centering\small
\caption{Effective benchmark denominators shared by all four backbones.}
\label{tab:app_benchmark_denominators}
\begin{tabularx}{\linewidth}{@{}lrX@{}}
\toprule
Benchmark & $n$ & Evaluation focus\\
\midrule
WorldSense & 3,172 & Integrated real-world audio--video understanding\\
Daily-Omni & 1,197 & Everyday audio--video understanding and reasoning\\
OmniVideoBench & 1,000 & Multitask audio--video understanding\\
LVOmniBench & 1,014 & Long audio--video understanding\\
AVUT-Human & 1,733 & Audio-centered video understanding\\
\bottomrule
\end{tabularx}
\end{table}

The released AV-Human split contains 1,734 questions; we exclude one item whose correct option is an empty string (QA\_id 455), and all methods and runs share the remaining 1,733 questions.

Long-video evaluation uses each model's frame budget. LVOmniBench uses the same preprocessed media for all compared objectives, with durations up to 2,100 seconds.

Within each backbone and benchmark, all methods and runs share the question set and denominator. For the five repeated configurations, means and sample SDs use unrounded accuracies (SD denominator 2). Task-equal macros are formed within runs: Avg-3 covers WorldSense, Daily-Omni, and AVUT-Human; Avg-5 covers all five tasks. Differences use unrounded means; rounding to two decimals occurs last. Base and Text CoT-SFT are single evaluations without training-run SD. Diagnostic Daily-Omni scores use diagnostic checkpoints. A displayed 0.00 SD can reflect rounding. The 20 positive SyncRA-minus-Vanilla benchmark mean differences form a consistent pattern across the evaluated settings, and the two dense and two MoE backbones together probe portability across architectures and update scopes.

\subsection{Alignment-loss objectives and rationale}
\label{app:loss_ablation}
We compare positive-only cosine alignment, pairwise logistic ranking, and SyncRA's InfoNCE objective on Qwen3-Omni, Run 1, with Vanilla SFT as the answer-only baseline. The four methods in Table~\ref{tab:loss-ablation-main} start from the same Base weights and share the data split, media processing, trainable scope, answer supervision, and optimization settings in Appendix~\ref{app:optimization}. The three auxiliary objectives use the same block-36 interval pools and shared rank-64 projection initialization. Each method is trained independently from Base and selects its checkpoint by validation answer-token NLL.

Using the normalized features in Eq.~\eqref{app:eq_projection}, the cosine-only objective is
\begin{equation}
\mathcal L_{\mathrm{cos}}=\frac{1}{K\tau}\sum_{k=1}^{K}\left[1-(z_k^a)^\top z_k^v\right],
\qquad \mathcal L=\mathcal L_{\mathrm{answer}}+0.1\mathcal L_{\mathrm{cos}},\quad \tau=0.07.
\label{app:eq_cosine_loss}
\end{equation}
The $1/\tau$ factor matches the coefficient of the explicit positive-pair attraction term in InfoNCE. Cosine-only rewards each matching pair independently, whereas InfoNCE also makes it compete against other intervals from the same video. This competition directly trains the distinction between the corresponding moment and alternative times. Samples with $K<2$ use answer supervision alone.

For $K\ge2$, pairwise logistic ranking uses
\begin{equation}
\mathcal L_{\mathrm{rank}}
=
\frac{1}{2K(K-1)}
\sum_{i=1}^{K}\sum_{j\ne i}
\left[
\operatorname{softplus}(s_{ij}-s_{ii})
+
\operatorname{softplus}(s_{ji}-s_{ii})
\right],
\label{app:eq_ranking_loss}
\end{equation}
where $\operatorname{softplus}(x)=\log(1+\exp(x))$ is evaluated using a numerically stable implementation. The logits already include temperature $\tau=0.07$, with no additional temperature factor. All off-diagonal within-video candidates are retained, with the same positive pairs as InfoNCE. The training objective is $\mathcal L_{\mathrm{answer}}+0.1\mathcal L_{\mathrm{rank}}$. Ranking compares positive and negative scores independently, while InfoNCE jointly normalizes the candidate scores. The shared loss weight does not imply matched total gradient norms.

\paragraph{Why contrastive interval classification?}
InfoNCE was introduced in contrastive predictive coding for learning representations through future prediction~\citep{oord2018cpc}. Here, temporal co-occurrence defines positives, while within-video candidates reduce source-identity shortcuts. Mapping all intervals to the same unit vector yields $\mathcal L_{\mathrm{cos}}=0$ but $\mathcal L_{\mathrm{SyncRA}}=\log K$ for $K\ge2$, showing that positive attraction alone does not enforce interval discrimination. This distinction is consistent with the alignment--uniformity perspective~\citep{wang2020alignmentuniformity}. Both ranking and InfoNCE weight harder negatives more strongly, but InfoNCE couples these weights through joint normalization. Their relative effectiveness is assessed by the matched training comparison, rather than assumed from this distinction alone.

\paragraph{Results and uncertainty.}
AllFour and Native R@1 follow the protocols in Appendices~\ref{app:behavior_metrics} and~\ref{app:representation}; Daily-Omni uses the same 1,197 questions. Cosine-only and pairwise ranking reach 40.50\% and 46.25\% AllFour, compared with 62.50\% for InfoNCE (Table~\ref{tab:loss-ablation-main}). The SyncRA-minus-control differences are 22.00 and 16.25 percentage points, with paired 95\% source-video bootstrap intervals of $[17.75,26.25]$ and $[12.25,20.25]$, respectively. Daily-Omni accuracies differ by at most 0.75 percentage points among the three objectives. The intervals use 10,000 paired source-video bootstrap resamples, retaining all cues and versions from each source, and quantify source-sampling uncertainty for the fixed Run~1 checkpoints.

\subsection{Ported prior synchrony objectives}
\label{app:ported}
We re-implement the auxiliary losses of \citet{korbar2018cooperative} (AVTS), CAV-MAE Sync \citep{araujo2025cavmaesync}, and AVAF-Net \citep{zhao2025avafnet} from their papers and port them to Qwen3-Omni. Only the auxiliary loss changes. The data, answer loss, trainable parameters, supervision block (36), two-epoch budget, NLL checkpoint selection, and loss weight 0.1 are identical to SyncRA under the Run~1 protocol, with a single seed. The projection sizes and the CAV-MAE Sync temperature are not specified in the original papers; all other choices follow the source.

\paragraph{Korbar et al.}
A squared-distance margin loss with margin $\eta=0.99$ on $L_2$-normalized 64-dimensional projections. Negatives follow the original curriculum, compressed to 1,584 steps: all negatives come from other videos for the first 880 steps, and 25\% come from the same video afterwards.

\paragraph{CAV-MAE Sync.}
Frame-level InfoNCE with temperature 0.07 and a shared 64-dimensional projection. Negatives come from other videos, and the loss runs in the video-to-audio direction.

\paragraph{AVAF-Net.}
A temporal affinity consistency loss with one 64-dimensional projection per modality, computed over the affinity structure of all intervals within each video.

All three losses converge during training: from 1.05 to 0.25 for Korbar et al., from 4.42 to 0.12 for CAV-MAE Sync, and from 2.01 to 0.09 for AVAF-Net. Table~\ref{tab:app_ported} gives the complete comparison.

\begin{table}[htbp]
\centering\small
\setlength{\tabcolsep}{2.4pt}
\caption{\textbf{Same-setting comparison with prior synchrony objectives on Qwen3-Omni (Run~1 protocol).} The $\Delta$ columns are AllFour differences in percentage points with paired 95\% source-video bootstrap intervals. $P(\mathrm{all}\,|\,\mathrm{O})$ is AllFour divided by original-version accuracy. R@1 is unprojected same-interval retrieval at block 36; values for the ported objectives use a refreshed pipeline (differs from Table~\ref{tab:representation} by $\le$0.1).}
\label{tab:app_ported}
\begin{tabular}{@{}lrllrrr@{}}
\toprule
Objective & AllFour & $\Delta$ vs.\ Vanilla SFT & $\Delta$ vs.\ SyncRA & $P(\mathrm{all}|\mathrm{O})$ & R@1 & Daily-Omni\\
\midrule
Vanilla SFT & 34.00 & \textemdash & $-28.50$ $[-33.25,-23.75]$ & 38.3 & 13.4 & 75.86\\
Korbar et al. & 33.25 & $-0.75$ $[-3.75,+2.25]$ & $-29.25$ $[-33.75,-24.75]$ & 37.8 & 12.1 & 77.27\\
CAV-MAE Sync & 52.50 & $+18.50$ $[13.50,23.50]$ & $-10.00$ $[-14.25,-5.75]$ & 57.2 & 53.6 & 78.36\\
AVAF-Net & 34.50 & $+0.50$ $[-2.75,+3.75]$ & $-28.00$ $[-32.25,-23.50]$ & 39.1 & 11.8 & 77.94\\
SyncRA & \textbf{62.50} & $+28.50$ $[23.75,33.25]$ & \textemdash & \textbf{67.0} & \textbf{96.9} & \textbf{78.95}\\
\bottomrule
\end{tabular}
\end{table}

For Korbar et al., between 75\% and 100\% of the negatives come from other videos, so the margin loss is mostly satisfied by telling videos apart, and within-video retrieval stays at the Vanilla level. The ported CAV-MAE Sync differs from SyncRA in the source of its negatives and in the direction of its loss, and it lands in the middle on both R@1 and AllFour. For AVAF-Net, the loss converges while within-video R@1 stays at 11.8\%.

\clearpage
\section{Audio--Visual Correspondence Selection}
\label{app:behavior}

\subsection{Source material and annotation}
The diagnostic uses one continuous clip from each of 100 independent source videos. Daily-Omni contributes 81 clips: 80 of 60 seconds and one of 53.5 seconds. LVOmniBench contributes 19 clips of 120 seconds. Each source supplies four audio cues and four input versions, giving 400 cues and 1,600 questions per checkpoint. The 100 diagnostic source IDs do not appear among the training or validation sources of this subset.

Each clip contains four selected equal-length, nonoverlapping windows, $W_1,\ldots,W_4$, with one complete multiword spoken phrase per window. Every phrase occurs exactly once in the evaluated clip and lies at least one second from either window boundary. The four scene descriptions are specific and mutually exclusive, and all describe content present in the clip. Throughout each phrase's playback, every input version must show a stable, unique target scene that is also visible in the frames actually sampled by Qwen3. Annotations combine independent ASR with human visual inspection of the finished media and are fixed before formal model inference.

\subsection{Input versions and question construction}
We apply the same exchange $W_1\leftrightarrow W_3$, $W_2\leftrightarrow W_4$ independently to audio and video. O retains both tracks, A exchanges audio, V exchanges video, and AV exchanges both. Each exchange moves every audio sample or video frame in the selected windows, preserving within-window order. Content outside the windows and total duration remain fixed. The original version passes through the same media-processing pipeline.

O and V share the same audio track, while A and AV share the exchanged audio track. O and A share the same video track, while V and AV share the exchanged video track. Table~\ref{tab:app_correspondence_keys} gives the resulting scene assignments. The four scene descriptions and their option order are fixed across all cues and versions of a source; scene-to-letter assignments are balanced across sources.

\begin{table}[htbp]
\centering\small
\caption{Correct scene for each audio cue across input versions. $X_i$ denotes the scene paired with cue $i$ in the original clip. Each source maps these scenes to a fixed order of option letters.}
\label{tab:app_correspondence_keys}
\begin{tabular}{@{}lcccc@{}}
\toprule
Cue & O & A & V & AV\\
\midrule
Cue 1 & $X_1$ & $X_3$ & $X_3$ & $X_1$\\
Cue 2 & $X_2$ & $X_4$ & $X_4$ & $X_2$\\
Cue 3 & $X_3$ & $X_1$ & $X_1$ & $X_3$\\
Cue 4 & $X_4$ & $X_2$ & $X_2$ & $X_4$\\
\bottomrule
\end{tabular}
\end{table}

Each question supplies the complete current audio--video clip and asks: ``While the phrase \{cue\} is being spoken, which of the following scenes is visible?'' The four lettered options are followed by ``Answer with A, B, C, or D only.'' The question contains no timestamp, window position, or version label. Questions are evaluated independently, without access to answers from other versions.

\paragraph{Example.}
The first source in the fixed diagnostic order is a 120-second LVOmniBench clip. Its four five-second windows are 13.5--18.5, 38--43, 60--65, and 109--114 seconds. One cue is ``until Christmas.'' The exact question and options are:
\begin{quote}\small
While the phrase ``until Christmas'' is being spoken, which of the following scenes is visible?\par
A. A bearded man wearing glasses, a blue-and-yellow cap and a dark plaid jacket speaks outside a log wall.\par
B. The camera looks across a wood-paneled living room, with a pale green sofa and a wooden table in the foreground.\par
C. The camera looks into a wood-paneled bedroom with a large white bed and dark wooden bunk beds.\par
D. A hand holds a smartphone outdoors over wooded ground; the screen displays a room photograph and a social-media post.\par
Answer with A, B, C, or D only.
\end{quote}
In O, the phrase plays at 15.41--16.44 seconds and the answer is A. In A, it moves to 61.91--62.94 seconds and the answer is C. In V, it stays at 15.41--16.44 seconds while the answer becomes C. In AV, it moves to 61.91--62.94 seconds and the answer remains A. These times describe the construction and are not given to the model.

\subsection{Checkpoints, metrics, and uncertainty}
\label{app:behavior_metrics}
We evaluate Base, Vanilla SFT, and SyncRA checkpoints for all
four backbones on the same frozen diagnostic.
Text CoT-SFT results are also reported.
The Qwen3-Omni comparisons additionally include Permuted pairs,
Fixed time codes, and Clip-level AV; its fine-tuned checkpoints
use Run~1.
Table~\ref{tab:attribution} reports AllFour for these three controls
on the other backbones.
Qwen2.5-Omni, MiniCPM-o 4.5, and Nemotron use their previously
selected epoch-2 fine-tuned checkpoints.
No checkpoint is reselected using this diagnostic.

Evaluation follows each backbone's native media processing,
chat template, and answer extraction.
MiniCPM uses 1 FPS and one-second audio chunks; the other
backbones use their existing 2 FPS processing paths.
Decoding is greedy, with a 1,280-token limit except for
MiniCPM Base, which retains its benchmark
\texttt{Final answer:} prefix and six-token limit.
Text CoT-SFT receives the same fixed letter-answer questions
without an additional reasoning prompt.
Generation and parsing failures remain in the denominator.

On Qwen3-Omni, the Text CoT-SFT checkpoint attains 12.00\% AllFour and 54.06\% overall accuracy. Its O/A/V/AV accuracies are 67.50/45.25/40.00/63.50\%, and its Daily-Omni accuracy is 74.44\%. This baseline evaluates supervision on supplied text reasoning; it is separate from the auxiliary-target comparison in Table~\ref{tab:objective-ablation}.

\begin{table}[htbp]
\centering
\small
\setlength{\tabcolsep}{4pt}
\caption{
\textbf{Complete correspondence-selection results.}
All entries are percentages.
AllFour requires correct answers in all four versions;
Overall averages all 1,600 question--version evaluations.
O/A/V/AV denote original, audio-exchanged, video-exchanged,
and jointly exchanged inputs.
}
\label{tab:correspondence-full}
\begin{tabular*}{\linewidth}
  {@{\extracolsep{\fill}}lrrrrrr@{}}
\toprule
Checkpoint & AllFour & Overall & O & A & V & AV \\
\midrule
\multicolumn{7}{@{}l}{\textit{Qwen2.5-Omni (dense)}} \\
Base         & 2.25  & 40.75 & 58.50 & 33.25 & 21.00 & 50.25 \\
Vanilla SFT  & 11.00 & 54.19 & 73.75 & 51.50 & 29.75 & 61.75 \\
Text CoT-SFT & 6.25  & 46.75 & 62.00 & 43.00 & 30.25 & 51.75 \\
SyncRA       & 20.75 & 64.31 & 77.75 & 62.75 & 46.25 & 70.50 \\
\midrule
\multicolumn{7}{@{}l}{\textit{MiniCPM-o 4.5 (dense)}} \\
Base         & 35.50 & 73.06 & 85.75 & 71.75 & 53.75 & 81.00 \\
Vanilla SFT  & 57.25 & 84.44 & 92.50 & 85.00 & 71.25 & 89.00 \\
Text CoT-SFT & 36.75 & 74.38 & 88.00 & 73.00 & 56.75 & 79.75 \\
SyncRA       & 62.25 & 86.50 & 93.00 & 87.00 & 75.75 & 90.25 \\
\midrule
\multicolumn{7}{@{}l}{\textit{Qwen3-Omni (MoE)}} \\
Base             & 13.25 & 56.38 & 76.50 & 47.25 & 30.75 & 71.00 \\
Vanilla SFT      & 34.00 & 72.25 & 88.75 & 67.00 & 53.00 & 80.25 \\
Text CoT-SFT     & 12.00 & 54.06 & 67.50 & 45.25 & 40.00 & 63.50 \\
Permuted pairs   & 31.00 & 70.94 & 86.75 & 65.25 & 51.25 & 80.50 \\
Fixed time codes & 44.75 & 80.19 & 91.75 & 77.50 & 64.00 & 87.50 \\
Clip-level AV    & 31.75 & 71.75 & 86.50 & 66.00 & 50.75 & 83.75 \\
SyncRA           & 62.50 & 85.50 & 93.25 & 81.00 & 76.50 & 91.25 \\
\midrule
\multicolumn{7}{@{}l}{\textit{Nemotron (MoE)}} \\
Base         & 0.50  & 36.00 & 52.00 & 31.25 & 17.00 & 43.75 \\
Vanilla SFT  & 6.00  & 51.00 & 74.25 & 48.50 & 27.75 & 53.50 \\
Text CoT-SFT & 4.50  & 44.06 & 58.75 & 39.25 & 28.50 & 49.75 \\
SyncRA       & 16.50 & 62.38 & 80.50 & 64.00 & 41.25 & 63.75 \\
\bottomrule
\end{tabular*}
\end{table}

Let $b_{iqv}$ indicate whether cue $q$ from source $i$ is answered correctly in version $v$. With $N=100$ sources, four cues per source, and $\mathcal V=\{\mathrm O,\mathrm A,\mathrm V,\mathrm{AV}\}$, the reported percentage metrics are
\begin{equation}
\begin{split}
\mathrm{AllFour}&=\frac{100}{4N}\sum_{i=1}^{N}\sum_{q=1}^{4}\prod_{v\in\mathcal V}b_{iqv},\\
\mathrm{Overall}&=\frac{100}{16N}\sum_{i=1}^{N}\sum_{q=1}^{4}\sum_{v\in\mathcal V}b_{iqv}.
\end{split}
\label{app:eq_correspondence_metrics}
\end{equation}
AllFour requires correctness in every version, rather than agreement among answers whose keys differ. Each source receives equal weight in both metrics. Condition-specific accuracies use the 400 questions for that version.

Paired bootstrap intervals use 10,000 resamples of the 100 source videos. Each selected source retains all four cues and all four versions, and methods are compared on identical resamples. The resulting 95\% intervals quantify variation across source videos with the evaluated checkpoints held fixed. Table~\ref{tab:app_cross_backbone_correspondence_deltas} reports the cross-backbone AllFour differences. Table~\ref{tab:app_correspondence_deltas} reports both aggregate metrics for the Qwen3-Omni objective comparisons, and Table~\ref{tab:app_correspondence_benchmarks} separates its source benchmarks.

\begin{table}[htbp]
\centering\small
\setlength{\tabcolsep}{8pt}
\caption{\textbf{AllFour improvements across backbones.}
SyncRA minus Vanilla SFT in percentage points on the same 100 source
videos. Paired 95\% source-video bootstrap intervals keep the diagnostic
checkpoints fixed; all four intervals exclude zero.}
\label{tab:app_cross_backbone_correspondence_deltas}
\begin{tabular}{@{}lcrr@{}}
\toprule
Backbone & Architecture & $\Delta$ & 95\% interval\\
\midrule
Qwen2.5-Omni  & Dense & $+9.75$  & $[5.50,14.00]$\\
MiniCPM-o 4.5 & Dense & $+5.00$  & $[1.00,9.00]$\\
Qwen3-Omni    & MoE   & $+28.50$ & $[23.75,33.25]$\\
Nemotron      & MoE   & $+10.50$ & $[6.75,14.50]$\\
\bottomrule
\end{tabular}
\end{table}

\begin{table}[htbp]
\centering\small
\setlength{\tabcolsep}{4.5pt}
\caption{Qwen3-Omni: SyncRA minus each comparator, in percentage points. Intervals are paired 95\% source-video bootstrap intervals. All checkpoints are fixed to Run 1.}
\label{tab:app_correspondence_deltas}
\begin{tabular}{@{}lrrrr@{}}
\toprule
& \multicolumn{2}{c}{AllFour} & \multicolumn{2}{c}{Overall accuracy}\\
Comparator & $\Delta$ & 95\% interval & $\Delta$ & 95\% interval\\
\midrule
Vanilla SFT & $+28.50$ & $[23.75,33.25]$ & $+13.25$ & $[11.19,15.38]$\\
Permuted pairs & $+31.50$ & $[26.75,36.25]$ & $+14.56$ & $[12.69,16.44]$\\
Fixed time codes & $+17.75$ & $[13.00,22.75]$ & $+5.31$ & $[3.44,7.19]$\\
Clip-level AV & $+30.75$ & $[26.25,35.00]$ & $+13.75$ & $[11.69,15.81]$\\
\bottomrule
\end{tabular}
\end{table}

\begin{table}[htbp]
\centering\small
\setlength{\tabcolsep}{5.5pt}
\caption{Qwen3-Omni correspondence selection by source benchmark (\%). LVOmniBench contributes 19 sources, 76 cues, and 304 questions per checkpoint; Daily-Omni contributes 81 sources, 324 cues, and 1,296 questions.}
\label{tab:app_correspondence_benchmarks}
\begin{tabular}{@{}lrrrr@{}}
\toprule
& \multicolumn{2}{c}{LVOmniBench} & \multicolumn{2}{c}{Daily-Omni}\\
Checkpoint & AllFour & Overall & AllFour & Overall\\
\midrule
Base & 7.89 & 50.00 & 14.51 & 57.87\\
Text CoT-SFT & 5.26 & 43.75 & 13.58 & 56.48\\
Vanilla SFT & 25.00 & 67.11 & 36.11 & 73.46\\
Permuted pairs & 28.95 & 67.76 & 31.48 & 71.68\\
Fixed time codes & 36.84 & 76.32 & 46.60 & 81.10\\
Clip-level AV & 27.63 & 69.41 & 32.72 & 72.30\\
SyncRA & \textbf{52.63} & \textbf{81.58} & \textbf{64.81} & \textbf{86.42}\\
\bottomrule
\end{tabular}
\end{table}

\subsection{Prediction matrices across input versions}
\label{app:prediction_matrices}
\begin{figure}[htbp]
\centering
\includegraphics[width=\linewidth]{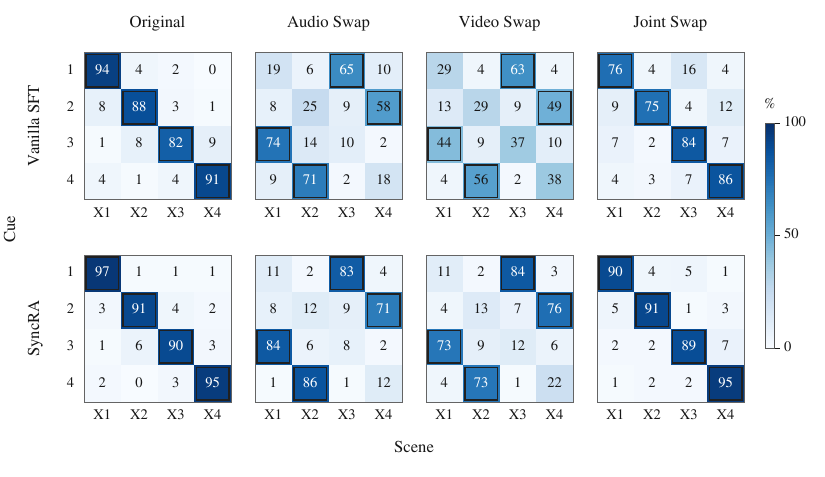}
\caption{Qwen3-Omni Run~1 scene-prediction matrices for Vanilla SFT and SyncRA. Each row aggregates one cue index over 100 source videos and sums to 100\%. Columns denote original scene identities $X_1,\ldots,X_4$, with option letters mapped back to these identities. Outlined cells mark the correct scene: the diagonal for original and joint-swap inputs, and the exchanged pairing ($1\leftrightarrow3$, $2\leftrightarrow4$) for audio- and video-swap inputs.}
\label{fig:app_prediction_matrices}
\end{figure}

Figure~\ref{fig:app_prediction_matrices} shows how scene predictions change with the audio--visual pairing. Under video exchange, the fraction of predictions retaining the original scene falls from 33.25\% for Vanilla SFT to 14.50\% for SyncRA, while accuracy rises from 53.00\% to 76.50\%. SyncRA also improves audio-exchange accuracy from 67.00\% to 81.00\%. When both modalities move together, predictions concentrate again on the original scene pairing, reaching 91.25\% accuracy compared with Vanilla's 80.25\%. These patterns show that SyncRA's answers follow the correspondence in the current input more reliably across the four versions.

\subsection{Control objectives and gain attribution}
\label{app:attribution}
Table~\ref{tab:attribution} reports the three auxiliary controls on all four backbones. The share captured by a control is its gain over Vanilla SFT divided by SyncRA's gain over Vanilla SFT. Avg-5 intervals are Welch 95\% intervals over the three training runs, and AllFour intervals are paired source-video bootstrap intervals as in Appendix~\ref{app:behavior_metrics}. Table~\ref{tab:app_attribution_shares} lists the shares.

\begin{table}[htbp]
\centering\small
\setlength{\tabcolsep}{6pt}
\caption{\textbf{Share of SyncRA's gain captured by the auxiliary controls.} The Avg-5 share is the range over the three controls; the AllFour share is that of the strongest control, Fixed time codes.}
\label{tab:app_attribution_shares}
\begin{tabular}{@{}lrlll@{}}
\toprule
& \multicolumn{3}{c}{Avg-5} & AllFour\\
\cmidrule(lr){2-4}\cmidrule(l){5-5}
Backbone & SyncRA $-$ Vanilla & 95\% interval & Control share & Control share\\
\midrule
Qwen2.5-Omni  & $+2.27$ & $[2.00,2.53]$ & $\le$25\% & 36\%\\
MiniCPM-o 4.5 & $+0.96$ & $[0.57,1.34]$ & controls below Vanilla SFT & 35\%\\
Qwen3-Omni    & $+1.93$ & $[1.65,2.21]$ & 46--73\% & 38\%\\
Nemotron      & $+2.11$ & $[1.88,2.34]$ & 35--45\% & 40\%\\
\bottomrule
\end{tabular}
\end{table}

On Avg-5, the share of the strongest control ranges from below zero to 73\%. On AllFour, it stays at 35--40\% on all four backbones, and the two controls without a correct temporal target, Permuted pairs and Clip-level AV, stay within $-3.00$ to $+1.50$ points of Vanilla SFT. On MiniCPM-o 4.5, SyncRA's AllFour advantage has an interval containing zero only against Fixed time codes ($+3.25$, $[-0.50,+7.00]$); the intervals against Vanilla SFT ($+5.00$, $[1.00,9.00]$), Clip-level AV ($+6.00$, $[2.25,9.75]$), and Permuted pairs ($+6.75$, $[2.75,10.75]$) exclude zero. The Base model of MiniCPM-o 4.5 already reaches 35.50\% AllFour and Vanilla SFT reaches 57.25\%, which leaves less headroom.

The five public benchmarks do not label whether a question requires audio--visual pairing, so the diagnostic is where the pairing-specific gain is measured. Every cue requires pairing by construction, and the gain holds for both cue types (Table~\ref{tab:app_filter_cues}).

\subsection{Robustness of the AllFour advantage}
\label{app:robustness}

\paragraph{Conditioning on original-version accuracy.}
We take the per-example predictions behind Table~\ref{tab:correspondence-full} and keep the cues that both Vanilla SFT and SyncRA answer correctly on the original version. On this set, we compare the joint accuracy on the A, V, and AV versions ($\Delta$joint; Table~\ref{tab:app_conditioning}). The advantage is unchanged on Qwen3-Omni and MiniCPM-o 4.5, and larger on Qwen2.5-Omni and Nemotron. On Qwen3-Omni, $\Delta$joint against the three auxiliary controls ranges from $+17.75$ to $+31.49$, and all intervals exclude zero. On the same backbone, AllFour divided by original-version accuracy rises from 38.3\% for Vanilla SFT to 67.0\% for SyncRA.

\begin{table}[htbp]
\centering\small
\setlength{\tabcolsep}{8pt}
\caption{\textbf{AllFour advantage after conditioning on original-version accuracy.} SyncRA minus Vanilla SFT in percentage points. $\Delta$joint is computed on the cues that both models answer correctly on the original version, with paired 95\% source-video bootstrap intervals.}
\label{tab:app_conditioning}
\begin{tabular}{@{}lrrrr@{}}
\toprule
Backbone & Raw $\Delta$AllFour & $\Delta$joint & 95\% interval & Joint set size\\
\midrule
Qwen2.5-Omni  & $+9.75$  & $+14.02$ & $[8.21,19.84]$  & 264\\
MiniCPM-o 4.5 & $+5.00$  & $+4.95$  & $[0.85,9.10]$   & 364\\
Qwen3-Omni    & $+28.50$ & $+28.86$ & $[23.68,34.10]$ & 350\\
Nemotron      & $+10.50$ & $+14.03$ & $[9.19,19.03]$  & 278\\
\bottomrule
\end{tabular}
\end{table}

\paragraph{Open-ended answers.}
We remove the options and ask each of the six Qwen3-Omni diagnostic checkpoints to describe the scene visible while the cue is spoken, which gives 1,600 answers per checkpoint. Two text-only LLM judges (Qwen3-Omni and Qwen2.5-Omni) independently match each answer to one of the four reference scenes or to ``no correspondence'', without knowing which checkpoint produced it. The two judges agree on 86.6\% of the answers ($\kappa=0.83$). Open-ended scores are lower than multiple-choice scores for every model, and SyncRA remains the highest (Table~\ref{tab:app_open_ended}). It exceeds the strongest control, Fixed time codes, under all three judging criteria: $+14.50$ $[9.50,19.50]$, $+15.50$ $[10.25,20.75]$, and $+11.50$ $[6.50,16.50]$. Under the Qwen3-Omni judge, AllFour divided by original-version accuracy is 43.4\% for SyncRA, 27.6\% for Fixed time codes, and 18.9\% for Vanilla SFT.

\begin{table}[htbp]
\centering\small
\setlength{\tabcolsep}{6pt}
\caption{\textbf{Open-ended swap test on Qwen3-Omni.} AllFour (\%) when the options are removed and an LLM judge matches each answer to a reference scene. ``Both'' requires both judges to accept all four versions.}
\label{tab:app_open_ended}
\begin{tabular}{@{}lrrrr@{}}
\toprule
& \multicolumn{3}{c}{Open-ended AllFour} & Multiple-choice\\
\cmidrule(lr){2-4}
Checkpoint & Qwen3-Omni judge & Qwen2.5-Omni judge & Both & AllFour\\
\midrule
Base             & 3.75  & 3.75  & 3.25  & 13.25\\
Vanilla SFT      & 13.00 & 11.75 & 9.25  & 34.00\\
Permuted pairs   & 14.00 & 13.50 & 11.00 & 31.00\\
Fixed time codes & 19.25 & 18.25 & 15.25 & 44.75\\
Clip-level AV    & 15.75 & 14.25 & 11.25 & 31.75\\
SyncRA           & \textbf{33.75} & \textbf{33.75} & \textbf{26.75} & \textbf{62.50}\\
\bottomrule
\end{tabular}
\end{table}

To check the judges, an author blind-labeled 200 answers: 100 random answers and 100 on which the judges disagree. The author agrees with the two judges on 78\% and 80\% of these answers. A judge chose ``no correspondence'' where the author recognized the correct scene in 9--12 answers, and a judge accepted an answer that the author rejected in 1--2 answers. The judges are therefore stricter than the author.

\paragraph{Larger answer-only budgets and data.}
We train Vanilla SFT and SyncRA on Qwen3-Omni for 2, 4, and 8 epochs on the 10K corpus, at about 60, 120, and 240 GPU-hours, and for 2 epochs on about 30K questions from about 4,000 videos. The runs are independent and use a single seed. The learning-rate schedule is rescaled to the total steps of each run, and checkpoints are selected by per-epoch validation NLL as in Appendix~\ref{app:optimization}. With 4$\times$ optimization steps or 3$\times$ data, Vanilla SFT stays at 34.00--36.75\% AllFour and the gap holds at $+27.25$ to $+28.50$ points (Table~\ref{tab:app_budget}). With 3$\times$ data, SyncRA's advantage changes by $-1.00$ $[-5.25,+3.25]$.

\begin{table}[htbp]
\centering\small
\setlength{\tabcolsep}{4.5pt}
\caption{\textbf{Larger answer-only budgets and data on Qwen3-Omni.} All checkpoints are NLL-selected. $\Delta$ is the AllFour difference, SyncRA minus Vanilla SFT, with paired 95\% source-video bootstrap intervals. R@1 is unprojected same-interval retrieval at block 36; values for the new runs use a refreshed pipeline (differs from Table~\ref{tab:representation} by $\le$0.1). The last column is the Daily-Omni difference.}
\label{tab:app_budget}
\begin{tabular}{@{}lrrrlrrr@{}}
\toprule
& \multicolumn{4}{c}{AllFour} & \multicolumn{2}{c}{R@1} & Daily-Omni\\
\cmidrule(lr){2-5}\cmidrule(lr){6-7}
Run & Vanilla & SyncRA & $\Delta$ & 95\% interval & Vanilla & SyncRA & $\Delta$\\
\midrule
10K, 2 epochs (main) & 34.00 & 62.50 & $+28.50$ & $[23.75,33.25]$ & 13.4 & 96.9 & $+3.09$\\
10K, 4 epochs & 34.75 & 62.75 & $+28.00$ & $[23.50,32.25]$ & 13.6 & 96.6 & $+3.01$\\
10K, 8 epochs & 35.00 & 62.25 & $+27.25$ & $[22.25,31.75]$ & 13.8 & 96.4 & $+2.92$\\
30K, 2 epochs & 36.75 & 64.25 & $+27.50$ & $[23.00,32.25]$ & 14.8 & 97.2 & $+3.01$\\
\bottomrule
\end{tabular}
\end{table}

Answer fitting saturates after the first epoch. The best validation answer NLL is 1.494, 1.492, and 1.491 for 2, 4, and 8 epochs, and the 8-epoch run overfits to 1.618 by its last epoch. SyncRA's validation answer NLL is slightly higher than Vanilla SFT's at every matched point (1.541 vs.\ 1.494 at 2 epochs), while its AllFour is far higher, so the gain does not come from better answer fitting. In the 8-epoch run, Vanilla SFT's best epoch reaches 35.75\% (epoch 4) and SyncRA's worst epoch reaches 55.75\% (epoch 1), and the gap is already $+24.50$ points after the first epoch.

\clearpage
\section{Additional Representation Analyses}
\label{app:representation}

\subsection{Cross-backbone heatmaps}
\label{app:cross-backbone-heatmaps}

We use the same two fixed validation clips across all backbones: Clip~1 (\texttt{79IElk5bfpY}) and Clip~2 (\texttt{U-vdQq3GOUw}).
Both clips were fixed before inspecting the additional
backbone results.
States are extracted from the original audio--video input,
question, and options, without target answers.
Within each native temporal interval, we mean-pool audio
and video states separately, apply L2 normalization, and
compute their cosine similarities without a projection.
Each backbone retains its native temporal grid.

\begin{figure}[p]
\centering
{\small Qwen3-Omni (MoE), block 36\par}
\includegraphics[width=\linewidth]{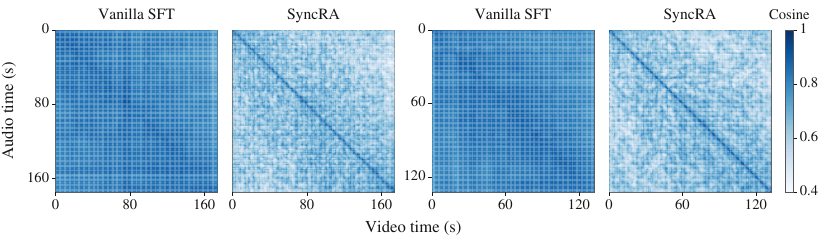}
\par\smallskip
{\small Qwen2.5-Omni (dense), block 21\par}
\includegraphics[width=\linewidth]{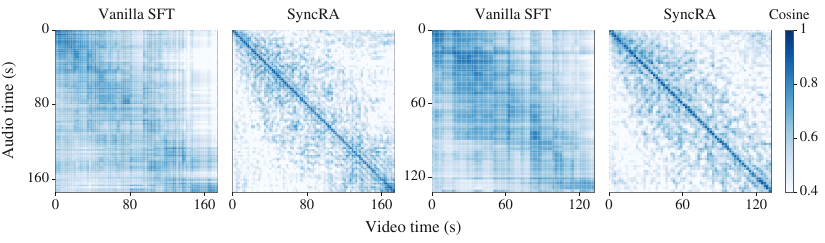}
\par\smallskip
{\small MiniCPM-o 4.5 (dense), block 9\par}
\includegraphics[width=\linewidth]{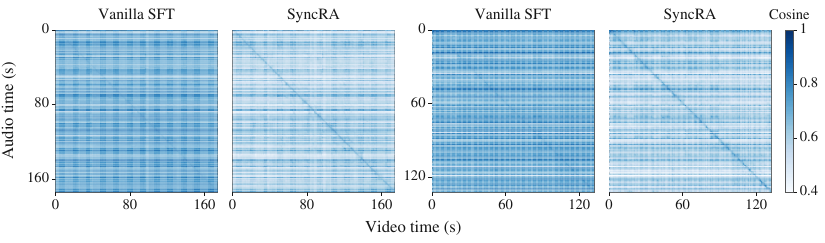}
\par\smallskip
{\small Nemotron (MoE), block 26\par}
\includegraphics[width=\linewidth]{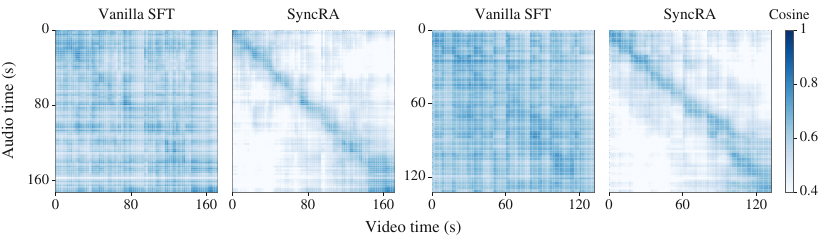}
\caption{
\textbf{Temporal similarity across four backbones.}
Each row shows Clip 1 followed by Clip 2, with Vanilla SFT
and SyncRA paired within each clip.
All panels use unprojected states and a shared color range
of 0.4--1; values below 0.4 saturate at the lightest color.
Axes show audio and video time in seconds.
}
\label{fig:cross-backbone-heatmaps}
\end{figure}

Figure~\ref{fig:cross-backbone-heatmaps} shows the resulting similarity matrices for all four backbones.

\subsection{Sample construction, states, and retrieval definitions}
The representation study uses Qwen3-Omni block 36 outputs, with hidden dimension 2,048. From the fixed validation split of this subset, we take the first 100 independent source videos and retain each source's first QA input. This yields 9,723 valid intervals in clips with annotated durations of 60--177 seconds. The modality states are extracted by prompt-only forward passes using identical complete audio--video inputs and the original question/options, without target answers or reasoning traces. Video uses 2 FPS with at most 256 frames, audio uses the full 16 kHz waveform, and the actual processor grid determines valid intervals.

This validation subset is drawn from the same development split whose parent set also serves answer-NLL checkpoint selection and layer selection on frozen Base states before fine-tuning. Each extraction uses its own available forward set. The representation analyses thus characterize the learned states on the development validation split. The time-decoder fitting set separately takes the first 100 independent training-source videos and their first QA inputs, yielding 10,249 intervals. Training and validation sources are disjoint. These representation diagnostics characterize temporal organization in jointly contextualized states.

For an evaluation video $i$ with $K_i$ intervals and normalized states $\hat a_{ik},\hat v_{ij}$, audio-to-video retrieval predicts
\begin{equation}
\hat j_i(k)=\arg\max_{j\in\{1,\ldots,K_i\}}\hat a_{ik}^{\top}\hat v_{ij},
\qquad R_i^{a\to v}=\frac{1}{|\mathcal Q_i|}\sum_{k\in\mathcal Q_i}\mathbf1[\hat j_i(k)=k].
\label{app:eq_retrieval}
\end{equation}
The video-to-audio definition is symmetric. We average the two directions within each video, then give all 100 videos equal weight:
\begin{equation}
R@1=\frac{1}{100}\sum_{i=1}^{100}\frac{R_i^{a\to v}+R_i^{v\to a}}{2}.
\label{app:eq_video_average}
\end{equation}
Retrieval time MAE replaces each hit indicator with the absolute difference between the predicted and target interval starting times and uses the same directional and video averaging. All-interval queries use $\mathcal Q_i=\{1,\ldots,K_i\}$. Video-equal chance is $100^{-1}\sum_i1/K_i\simeq1.11\%$.

Full-state retrieval normalizes the unprojected pooled state. The SyncRA-$W$ readout projects through the trained Run 1 SyncRA matrix and normalizes; exactly the same matrix is used for Vanilla and SyncRA. For the spectrum-matched random readout, take the thin SVD $W=U\Sigma V^\top$ and construct $R=U\Sigma Q^\top$, where $Q\in\mathbb R^{2048\times64}$ is the reduced-QR orthonormal basis of a standard Gaussian matrix generated with NumPy \texttt{default\_rng(1)}. This preserves $W$'s singular values while replacing its input subspace. One fixed $R$ is shared across all videos, both modalities, and both checkpoints. Retrieval normalizes the projected states, and Table~\ref{tab:representation} reports the resulting all-interval readouts.

\begin{table}[htbp]
\centering\small
\setlength{\tabcolsep}{6pt}
\caption{\textbf{Matching beyond the training projection.} Bidirectional same-interval retrieval, averaged equally over 100 validation videos. Each projected readout uses the same matrix for both checkpoints and modalities. MAE is measured in seconds.}
\label{tab:representation}
\begin{tabular}{@{}lrrrr@{}}
\toprule
& \multicolumn{2}{c}{R@1 (\%)} & \multicolumn{2}{c}{MAE (s)}\\
Space & Vanilla & SyncRA & Vanilla & SyncRA\\
\midrule
Unprojected states & 13.39 & 96.91 & 7.176 & 0.044\\
Trained SyncRA $W$ & 9.03 & 97.68 & 11.389 & 0.034\\
Spectrum-matched random & 7.58 & 80.72 & 15.332 & 1.334\\
\bottomrule
\end{tabular}
\end{table}

\begin{table}[t]
\centering\small
\setlength{\tabcolsep}{7pt}
\caption{\textbf{Loss-function ablation on Qwen3-Omni (Run 1).} Scores are percentages. Native R@1 uses unprojected block-36 states.}
\label{tab:loss-ablation-main}
\begin{tabular}{@{}lrrr@{}}
\toprule
Method & AllFour & Native R@1 & Daily-Omni\\
\midrule
Vanilla SFT & 34.00 & 13.39 & 75.86\\
Cosine-only & 40.50 & 17.92 & 78.20\\
Pairwise logistic ranking & 46.25 & 27.47 & 78.36\\
SyncRA (InfoNCE) & \textbf{62.50} & \textbf{96.91} & \textbf{78.95}\\
\bottomrule
\end{tabular}
\end{table}

\subsection{Audio replacement, delay, and the projection complement}
\label{app:audio_controls}
We modify the audio while preserving the video and current input positions, then extract states with a new forward pass. For donor replacement, sort the 100 validation sources by annotated clip duration, breaking ties by video ID, and exchange audio within each adjacent pair. This gives 50 pairs, uses each donor once, and fixes the same mapping for both checkpoints. Audio is copied from the donor waveform's start, with tail cropping or right zero-padding to preserve the recipient's waveform length.

The delay shifts audio by two intervals on each video's actual processor grid. With interval duration $\delta_i=2/f_i$ seconds, where $f_i$ is the actual video sampling FPS, the offset is rounded to the nearest integer number of samples. Prepending zeros and truncating the tail keeps the waveform length fixed. The resulting delays span approximately 2.00--2.77 seconds in this sample.

For all three audio conditions, the query set excludes two intervals at each boundary: zero-based indices $2,\ldots,K_i-3$. Candidates remain all $K_i$ intervals. The original-audio row therefore uses a different query denominator from the all-interval table. In the delay condition, current-interval targets have the same index as the query. Content-origin targets are $v_{k-2}$ for an audio query $a_k$ and $a_{k+2}$ for a visual query $v_k$.

For the complement readout, reduced QR of $W^\top$ gives an orthonormal row-space basis $Q_W\in\mathbb R^{2048\times64}$. We remove that component from each pooled state offline:
\begin{equation}
h_\perp=h-Q_WQ_W^\top h,
\label{app:eq_complement}
\end{equation}
and normalize the residual before retrieval. Its dimensionality is 1,984. This operation changes the diagnostic readout, without rerunning the model on altered hidden states.

\begin{table}[htbp]
\centering\small
\caption{Interior-query retrieval under audio replacement and delay. Every value is bidirectional, video-equal R@1 (\%). Queries exclude the first and last two native intervals; all intervals remain candidates. The complement removes the row-space component of the same SyncRA $W$ used for both checkpoints.}
\label{tab:app_interior_readout}
\begin{tabular}{llrrr}
\toprule
Checkpoint & Audio input & Unprojected states & SyncRA $W$ & $W$ complement\\
\midrule
Vanilla & Original & 12.75 & 8.35 & 12.69\\
Vanilla & Donor replacement & 9.28 & 6.10 & 9.10\\
Vanilla & Two-interval delay & 7.82 & 6.48 & 7.76\\
\midrule
SyncRA & Original & 97.07 & 97.78 & 96.87\\
SyncRA & Donor replacement & 96.74 & 97.41 & 96.59\\
SyncRA & Two-interval delay & 96.45 & 97.47 & 96.21\\
\bottomrule
\end{tabular}
\end{table}

\begin{table}[htbp]
\centering\small
\caption{Retrieval targets after a two-interval audio delay. Current-interval and content-origin targets are evaluated on the same interior queries and complete candidate set. Entries are R@1 (\%).}
\label{tab:app_delay_targets}
\begin{tabular}{lrrrr}
\toprule
& \multicolumn{2}{c}{Vanilla} & \multicolumn{2}{c}{SyncRA}\\
\cmidrule(lr){2-3}\cmidrule(lr){4-5}
Space & Current interval & Content origin & Current interval & Content origin\\
\midrule
Unprojected states & 7.82 & 7.86 & 96.45 & 0.01\\
SyncRA $W$ & 6.48 & 6.15 & 97.47 & 0.02\\
\bottomrule
\end{tabular}
\end{table}

After removing the 64-dimensional row-space component of $W$ from the pooled states, SyncRA retains 96.87\%, 96.59\%, and 96.21\% R@1 under original audio, donor replacement, and delay. The matching improvement therefore extends beyond that readout subspace.

Table~\ref{tab:app_control_retrieval} repeats the block-36 retrieval and the delay test for the three auxiliary controls under the same protocol. Without the delay, SyncRA exceeds every control by 74--84 points. After the delay, Fixed time codes shows the largest gap among the controls between the two targets (17.81\% vs.\ 11.53\%), and the other two differ by less than 0.5 points.

\begin{table}[htbp]
\centering\small
\caption{Block-36 retrieval for the auxiliary controls, without and with a two-interval audio delay. Entries are R@1 (\%) on unprojected states; the delay columns use the interior queries of Table~\ref{tab:app_delay_targets}. All rows use a refreshed pipeline (differs from Tables~\ref{tab:representation} and~\ref{tab:app_delay_targets} by $\le$0.1).}
\label{tab:app_control_retrieval}
\begin{tabular}{lrrrr}
\toprule
& & & \multicolumn{2}{c}{After delay}\\
\cmidrule(lr){4-5}
Checkpoint & AllFour & R@1 & Current interval & Content origin\\
\midrule
Vanilla SFT      & 34.00 & 13.48 & 7.92  & 7.79\\
Permuted pairs   & 31.00 & 12.74 & 7.85  & 7.57\\
Fixed time codes & 44.75 & 22.61 & 17.81 & 11.53\\
Clip-level AV    & 31.75 & 13.91 & 8.89  & 8.41\\
SyncRA           & 62.50 & 96.89 & 96.44 & 0.01\\
\bottomrule
\end{tabular}
\end{table}

\subsection{Linear decoding of elapsed time}
\label{app:linear_time}
For each checkpoint and source modality, we separately fit a ridge decoder to predict interval starting time from the block-36 state. Fitting uses the fixed 100-source training set (10,249 intervals), and evaluation uses the fixed 100-source validation set (9,723 intervals). Each training video receives equal total weight, distributed uniformly over its intervals. Features and time labels are centered with these weights, and the fit includes an intercept. Regularization is 0.001 times the mean eigenvalue of the weighted centered feature covariance. The solve is deterministic. MAE likewise averages intervals within each video, then averages videos. Cross-modal application keeps the source decoder's coefficients, training mean, and intercept unchanged.

The shuffled-label control independently permutes time labels within each training video, preserving that video's set of times; validation always uses the true labels. NumPy \texttt{default\_rng(20260909)} produces one set of permutations, with the same video order across checkpoints and the same shuffled labels for audio and video. This control uses a single shuffle realization.

Paired 95\% confidence intervals quantify evaluation-video sampling variation with the checkpoints and fitted decoders held fixed. We first compute each video's MAE difference for each compared pair (SyncRA minus Vanilla SFT, and Vanilla SFT minus Base). A separately initialized NumPy \texttt{default\_rng(20260909)} draws 100 validation videos with replacement in each of 2,000 resamples; all comparisons share these indices. We average the selected video differences without refitting the decoders and take the 2.5th and 97.5th percentiles using linear interpolation.

\begin{table}[htbp]
\centering\small
\caption{Elapsed-time ridge decoding. MAEs and differences are seconds; negative differences favor SyncRA. Differences are computed before rounding. Paired intervals resample evaluation videos while holding models and decoders fixed. Source-modal centering and intercept remain fixed under transfer. Base is the released checkpoint before fine-tuning; Vanilla-minus-Base differences are reported in the text.}
\label{tab:app_linear_time}
\begin{tabular}{lrrrrr}
\toprule
Fit $\to$ evaluate & Base MAE & Vanilla MAE & SyncRA MAE & Difference & Paired 95\% interval\\
\midrule
Audio $\to$ audio & 8.996 & 8.188 & 6.680 & $-1.508$ & $[-1.881,-1.168]$\\
Video $\to$ video & 5.675 & 5.157 & 4.651 & $-0.506$ & $[-0.851,-0.163]$\\
Audio $\to$ video & 13.133 & 12.685 & 15.741 & $+3.056$ & $[1.677,4.450]$\\
Video $\to$ audio & 18.107 & 18.051 & 17.290 & $-0.760$ & $[-1.484,-0.007]$\\
\bottomrule
\end{tabular}
\end{table}

\begin{table}[htbp]
\centering\small
\caption{Controls for elapsed-time decoding. Values are validation MAE in seconds under video-balanced evaluation. Time labels are shuffled once within each training video, with the same shuffled labels shared across modalities and checkpoints. The constant predictor uses the training-label mean.}
\label{tab:app_time_baselines}
\begin{tabular}{lrrr}
\toprule
Control & Base & Vanilla & SyncRA\\
\midrule
Shuffled time labels: audio $\to$ audio & 30.816 & 30.249 & 29.617\\
Shuffled time labels: video $\to$ video & 31.041 & 30.196 & 29.044\\
Shuffled time labels: audio $\to$ video & 39.298 & 54.523 & 30.236\\
Shuffled time labels: video $\to$ audio & 60.822 & 55.117 & 62.424\\
Training-label-mean constant & 29.276 & 29.276 & 29.276\\
\bottomrule
\end{tabular}
\end{table}

The released Base checkpoint already exposes linearly recoverable time: its within-modality MAEs of 8.996 seconds (audio) and 5.675 seconds (video) sit far below its shuffled-label controls (30.816 and 31.041 seconds) and the 29.276-second constant predictor. Answer-only fine-tuning sharpens this within-modality readout rather than creating it: Vanilla-minus-Base differences are $-0.808$ seconds $[-1.053,-0.571]$ for audio and $-0.518$ seconds $[-0.761,-0.278]$ for video, while cross-modal transfer does not change significantly ($-0.448$ seconds $[-1.180,+0.261]$ audio to video; $-0.056$ seconds $[-0.576,+0.458]$ video to audio). SyncRA reduces within-modality MAE further, while direct transfer changes differently by direction, including worse audio-to-video transfer. Recovering scalar time and matching intervals across modalities measure distinct properties: high same-interval retrieval need not yield a time decoder that transfers well across modalities.

\subsection{Clip-level retrieval across source videos}
\label{app:clip_retrieval}
We compare Vanilla, SyncRA, and Clip-level AV Run 1 checkpoints on the same 100 OmniVideo validation videos and block-36 states. For each modality, mean-pool its valid interval states, then apply the specified projection and L2 normalization before cosine retrieval of the matching source video. The full pool contains all 100 sources (chance 1\%). A duration-neighbor pool contains the correct source and the nine other videos with closest annotated duration, with ties broken by fixed source order (chance 10\%). Directions receive equal weight. Each named $W$ is held identical across all three checkpoints.

\begin{table}[htbp]
\centering\small
\caption{Clip-level cross-video retrieval, R@1 (\%). Both candidate pools use the same 100 validation sources. The reported $W$ is shared across checkpoints, including when its training objective differs from the evaluated checkpoint's objective.}
\label{tab:app_clip_retrieval}
\begin{tabular}{llrrr}
\toprule
Candidate pool & Space & Vanilla & SyncRA & Clip-level AV\\
\midrule
All 100 sources & Unprojected states & 8.50 & 15.00 & 18.50\\
All 100 sources & Clip-level AV $W$ & 5.50 & 36.50 & 19.00\\
All 100 sources & SyncRA $W$ & 7.50 & 68.50 & 20.50\\
\midrule
10 duration neighbors & Unprojected states & 34.50 & 41.50 & 50.00\\
10 duration neighbors & Clip-level AV $W$ & 30.00 & 65.00 & 54.00\\
10 duration neighbors & SyncRA $W$ & 31.50 & 86.00 & 57.50\\
\bottomrule
\end{tabular}
\end{table}

The ordering across objectives depends on both readout space and temporal granularity. Clip-level AV yields the highest pooled full-state retrieval, while SyncRA is strongest through either learned projection. These clip-level results complement within-video interval retrieval rather than replacing its measurement of temporal correspondence. All features here are contextualized internal states extracted after joint audio--video input, rather than independent unimodal encoders.

\clearpage
\subsection{Correspondence across model depth}
\label{app:depth_analysis}
\label{app:depth_scan}
We extract block outputs at depths 1, 12, 24, 36, and 48 from fixed Qwen3-Omni Base, Vanilla SFT, and SyncRA checkpoints. The scan uses the same 100 validation videos and 9,723 valid intervals as the block-36 study in Appendix~\ref{app:representation}. At every depth, retrieval uses L2-normalized, unprojected interval-mean states, with all valid intervals as queries and candidates. R@1 and retrieval-time MAE average the two directions within each video before averaging videos equally; the chance level is 1.11\%.

SyncRA substantially strengthens correspondence in the intermediate and later representations. At block 24, R@1 increases from Vanilla's 7.83\% to 19.96\%. At the supervised block 36, it rises from 13.39\% to 96.91\%, while retrieval-time MAE drops from 7.176 to 0.044 seconds. The final block also retains a clear advantage, reaching 31.06\% versus Vanilla's 3.44\%. Together with the frozen-Base probe scores in Table~\ref{tab:candidates}, this profile supports placing temporal supervision at an intermediate layer where cross-modal correspondence is accessible and can be substantially strengthened during QA training.

Table~\ref{tab:app_final_layer} adds the three auxiliary controls at the final block under the same protocol. SyncRA's R@1 exceeds Fixed time codes by 23.58 points $[22.27,24.87]$ and Vanilla SFT by 27.63 points $[26.28,28.98]$, with paired 95\% per-video bootstrap intervals from 10,000 resamples. Fixed time codes matches SyncRA on retrieval-time MAE (difference $+0.02$~s, $[-1.13,+1.18]$) but retrieves the exact interval at 7.49\%: it brings the retrieved interval close in time, and rarely to the exact interval.

\begin{table}[htbp]
\centering\small
\setlength{\tabcolsep}{8pt}
\caption{\textbf{Final-layer retrieval on Qwen3-Omni.} Bidirectional same-interval retrieval on unprojected block-48 states, averaged equally over 100 validation videos. Rows for the controls use a refreshed pipeline (differs from Table~\ref{tab:representation} by $\le$0.1).}
\label{tab:app_final_layer}
\begin{tabular}{@{}lrr@{}}
\toprule
Checkpoint & R@1 (\%) & Retrieval-time MAE (s)\\
\midrule
Base             & 3.22  & 30.17\\
Vanilla SFT      & 3.44  & 29.61\\
Permuted pairs   & 3.40  & 29.18\\
Fixed time codes & 7.49  & 14.24\\
Clip-level AV    & 3.68  & 27.81\\
SyncRA           & \textbf{31.06} & 14.26\\
\bottomrule
\end{tabular}
\end{table}

\paragraph{Supervision-depth ablation.}
We compare supervision at blocks 24 and 36 of Qwen3-Omni
using seed 1 (Run~1), starting from the same Base weights
and identical shared rank-64 projection initialization.
Both runs use the same data and optimization settings,
with $\lambda=0.1$ and $\tau=0.07$, and train for two epochs
(1,584 steps). Checkpoints are selected by validation
answer-token NLL; both selected checkpoints are from epoch 2.
AllFour uses the 100-source-video diagnostic in
Appendix~\ref{app:behavior}, and Daily-Omni uses the same
1,197 evaluation questions. Native R@1 is measured at
both blocks on the same 100 validation videos, following
the unprojected retrieval and averaging protocol above.

\begin{table}[htbp]
\centering
\small
\setlength{\tabcolsep}{6pt}
\caption{\textbf{Supervision-depth ablation on Qwen3-Omni (Run~1).}
Scores are percentages. Native R@1 uses unprojected states at the
indicated readout block. The AllFour difference (block 36 minus
block 24) is +2.50 pp, with a paired 95\% source-video bootstrap
interval of $[-2.50,+7.75]$.
Both configurations use the training and evaluation protocol described above.}
\label{tab:supervision-depth}
\begin{tabular}{@{}lrrrr@{}}
\toprule
Supervision & \multicolumn{2}{c}{Native R@1}
            & AllFour & Daily-Omni\\
\cmidrule(lr){2-3}
block & Block 24 & Block 36 & & \\
\midrule
24           & 68.37 & 39.87 & 60.00 & 77.36\\
36 (default) & 19.96 & 96.91 & 62.50 & 78.95\\
\bottomrule
\end{tabular}
\end{table}

Table~\ref{tab:supervision-depth} shows that both tested supervision blocks improve AllFour over Vanilla SFT's 34.00\%. Supervision at block 24 produces stronger native correspondence at block 24, while supervision at block 36 produces stronger correspondence at block 36. The default block-36 configuration has higher point estimates for AllFour and Daily-Omni. The AllFour difference between the two locations has a paired interval containing zero, so this comparison supports the effectiveness of both tested intermediate locations and documents how placement shapes the depth-wise distribution of correspondence.

\paragraph{Auxiliary-weight ablation.}
\label{app:aux_weight}
We vary $\lambda\in\{0.03,0.1,0.3\}$ on Qwen3-Omni with seed 1, holding all other settings identical to Run 1 (Appendix~\ref{app:optimization}); checkpoints are selected by validation answer-token NLL. All three values improve AllFour over Vanilla SFT's 34.00\% by at least 24.00 points (Table~\ref{tab:lambda}).

\begin{table}[htbp]
\centering\small
\caption{\textbf{Auxiliary-weight ablation on Qwen3-Omni (seed 1).}
Native R@1 uses unprojected block-36 states on the 100 validation videos.
Paired 95\% source-video bootstrap intervals for the AllFour advantage of
$\lambda=0.1$ over $\lambda=0.03$ and $\lambda=0.3$ are $[+1.00,+8.00]$
and $[+0.25,+7.25]$ points, both excluding zero.}
\label{tab:lambda}
\begin{tabular}{@{}lrrr@{}}
\toprule
$\lambda$ & Native R@1 & AllFour & Daily-Omni\\
\midrule
0.03           & 90.85 & 58.00 & 78.53\\
0.1 (default)  & \textbf{96.91} & \textbf{62.50} & \textbf{78.95}\\
0.3            & 93.72 & 58.75 & 78.36\\
\bottomrule
\end{tabular}
\end{table}

\subsection{Activation patching}
\label{app:patching}
We run Qwen3-Omni on two inputs: a clip from the swap test, and the same clip with its video windows exchanged (Appendix~\ref{app:behavior}). The audio is identical in the two inputs, but a different scene is shown while the cue is spoken, so the correct answer is different. We then copy a chosen subset of hidden states from the second run into the first run, from block 4 to the last block, and measure how far the answer moves toward the answer of the second input. An effect of 0 means that the answer does not move, and an effect of 1 means that it moves as far as when the model is run on the exchanged input itself.

We call the window in which the cue is spoken the cue window, and the window exchanged with it the partner window. We use 400 patching instances (50 source videos $\times$ 4 cues $\times$ 2 input pairs) and analyze the instances that each checkpoint answers correctly without patching: 215 for Vanilla SFT and 289 for SyncRA. Both checkpoints are from Run~1.

\begin{table}[htbp]
\centering\small
\setlength{\tabcolsep}{6pt}
\caption{\textbf{Activation patching on Qwen3-Omni.} Normalized effect of copying hidden states from the video-exchanged input, from block 4 to the last block. The last two columns give SyncRA minus Vanilla SFT with its 95\% interval.}
\label{tab:app_patching}
\begin{tabular}{@{}lrrrr@{}}
\toprule
Hidden states copied from the exchanged input & Vanilla SFT & SyncRA & $\Delta$ & 95\% interval\\
\midrule
Video tokens of the cue and partner windows & 0.99 & 0.99 & \textemdash & \textemdash\\
Video tokens of the cue window only         & 0.91 & 0.97 & $+0.06$ & $[0.03,0.08]$\\
Video tokens of the partner window only     & 0.12 & 0.02 & $-0.10$ & $[-0.13,-0.07]$\\
Audio tokens of the cue window only         & 0.32 & 0.46 & $+0.14$ & $[0.11,0.17]$\\
\bottomrule
\end{tabular}
\end{table}

The answer depends on the video states of the cue window (Table~\ref{tab:app_patching}). Copying only these states flips the answer in 93\% of the instances for SyncRA and in 83\% for Vanilla SFT. SyncRA depends less on the partner window, with an effect of 0.02 against 0.12 for Vanilla SFT. The audio input is the same in the two runs, so the audio states of the cue window differ only through what they have taken from the video; copying them has a larger effect in SyncRA.

The two checkpoints read these states at the same depth. When the copy starts at block 24 instead of block 4, about half of the effect remains (0.53 for Vanilla SFT and 0.52 for SyncRA). When it starts at block 36, the effect is about 0 for both. The answer therefore reads the cue-window states before block 36. SyncRA applies its loss at the output of block 36, so its gradient shapes the blocks in which this reading happens.

\section{More Discussion}
\label{app:discussion}

\subsection{Comparison with LatentOmni}
\label{app:latentomni}
LatentOmni \citep{dai2026latentomni} is the closest concurrent work: it trains a temporal audio--visual contrastive signal jointly with interleaved text--latent reasoning and aligns extra latent states that it generates at inference, while SyncRA aligns the answering model's own intermediate states during QA fine-tuning and leaves inference unchanged. A controlled comparison on our testbeds requires running LatentOmni under the same backbones, data, and budget, which in turn requires its training and inference code, model checkpoints, and the LatentOmni-Instruct-35K dataset. As of November 17, 2026 (commit \texttt{7b2cb4e}), the official repository contains training and inference code for Qwen2.5-Omni only, the model weights are unreleased, and the dataset page is empty. The loss also cannot be ported on its own: the official \texttt{temporal\_sync\_loss} is defined on last-layer hidden states at the generated latent positions and is computed only during the latent-reasoning segment, so it depends on the full latent-reasoning pipeline and does not fit the answer-only setting of our comparisons. Our controlled comparisons therefore cover the training objectives in Section~\ref{sec:setup} and the three ported synchrony objectives in Section~\ref{sec:loss_ablation}, which share data, budget, and backbone.

\subsection{Training-data scale}
\label{app:data_scale}
The main fine-tuning experiments use the same compact 10K-question corpus (Appendix~\ref{app:training}). The corpus size is a deliberate design choice: every compared objective trains on identical data with an identical budget, so the benchmark differences isolate the supervision signal rather than data scale or annotation quality. SyncRA's interval labels come from the processor's existing timing metadata, and the recipe extends to larger corpora with no annotation cost. The AllFour advantage is not specific to this corpus size: on Qwen3-Omni it is $+27.50$ points with about 30K questions, against $+28.50$ points in the main setting (Appendix~\ref{app:robustness}). Behavior beyond 30K questions remains open (Appendix~\ref{sec:limitations}).

\subsection{Number of training runs}
\label{app:run_count}
Each trained configuration has three independent runs, and the summaries keep run-to-run variation visible: means use unrounded accuracies, sample SDs use denominator two, and task-equal macros are formed within runs before averaging (Appendix~\ref{app:benchmark_protocol}). The statistical evidence for SyncRA is the consistency of the paired differences rather than the precision of any single run. SyncRA exceeds Vanilla SFT in all 20 model--benchmark combinations under shared question sets and denominators, and on every backbone the Avg-5 gain far exceeds the run-to-run variation of either method (Section~\ref{sec:benchmarks}). On the diagnostic, the four cross-backbone paired 95\% source-video bootstrap intervals all exclude zero (Appendix~\ref{app:behavior_metrics}). Three runs per configuration reflect the compute needed to keep five training objectives comparable across four backbones; the 20-combination pattern, not any single run, carries the benchmark conclusion, and the pairing-specific conclusions rest on the diagnostic.

\subsection{Sync judgment under audio shifts}
\label{app:sync_judgment}
We test explicit sync judgment in the desynchronized case. We take 176 talking-face clips from WorldSense \citep{hong2026worldsense} and make four versions of each: two in sync and two with the audio shifted by $+2$~s and $-2$~s. We ask ``Is the audio in sync with the video?'' and read the result in two ways. The first is the generated answer, scored by balanced accuracy. The second is a threshold-free logit readout that compares $\log P(\text{out-of-sync})-\log P(\text{in-sync})$ across the versions of the same clip.

\begin{table}[htbp]
\centering\small
\setlength{\tabcolsep}{7pt}
\caption{\textbf{Sync judgment on Qwen3-Omni checkpoints.} 176 clips with two in-sync and two $\pm2$~s shifted versions each. Balanced accuracy and the logit readout are in \%.}
\label{tab:app_sync}
\begin{tabular}{@{}lrrrr@{}}
\toprule
Checkpoint & Balanced accuracy & Answers ``in sync'' & Logit readout & 95\% interval\\
\midrule
Base             & 49.4 & 97.3\% & 51.1 & $[47.3,55.0]$\\
Vanilla SFT      & 50.0 & 98.9\% & 51.1 & $[47.2,55.0]$\\
Permuted pairs   & 49.9 & 99.0\% & 53.0 & $[49.0,56.9]$\\
Fixed time codes & 50.1 & 97.0\% & 55.5 & $[51.8,59.2]$\\
Clip-level AV    & 49.6 & 97.6\% & 49.9 & $[46.1,53.7]$\\
SyncRA           & 49.7 & 97.4\% & 55.0 & $[51.1,58.7]$\\
\bottomrule
\end{tabular}
\end{table}

Every checkpoint, including Base, answers ``in sync'' on 97\% to 99\% of the inputs, and balanced accuracy stays between 49.4\% and 50.1\% (Table~\ref{tab:app_sync}). On the logit readout, SyncRA minus Vanilla SFT is $+3.9$ $[-1.6,+9.4]$. The test has a floor effect: no checkpoint detects a 2~s shift in its generated answers, so the test shows no measurable change from SyncRA and cannot tell whether the ability is preserved or reduced. In a spot check of 30 clips, 80\% show a clearly visible speaker. Explicit sync judgment is a separate capability from input-time binding.

\subsection{Training-interval audit and filtering ablation}
\label{app:interval_audit}

\paragraph{Audit.}
We sample 500 training intervals, 5 from each of 100 videos, take a 3~s clip for each, and label the clips automatically with Qwen3-Omni and Whisper \citep{radford2023whisper}. The two tools agree on speech presence in 90.4\% of the intervals. An author blind-labels 100 of the intervals, one per video, seeing only the 3~s clip.

\begin{table}[htbp]
\centering\small
\setlength{\tabcolsep}{8pt}
\caption{\textbf{Audit of the supervised training intervals.} Share of intervals in each category, with 95\% intervals for the two main categories.}
\label{tab:app_audit}
\begin{tabular}{@{}lll@{}}
\toprule
Category & Automatic labels (500) & Author blind labels (100)\\
\midrule
Visible sound source & 57.2\% $[49.6,64.8]$ & 40\% $[30,50]$\\
Off-screen narration or voice-over & 34.6\% $[27.6,41.6]$ & 55\% $[45,65]$\\
Music & 5.8\% & 4\%\\
Unrelated & 2.4\% & 1\%\\
\bottomrule
\end{tabular}
\end{table}

Intervals with no audio--visual relation are rare, and intervals with no visible sound source are common: 43\% by automatic labels and 60\% by author labels (Table~\ref{tab:app_audit}). The author and the automatic labels agree on 78\% of the intervals ($\kappa=0.62$). The main disagreement is narration over matching footage, which the automatic labeler marks as a visible source in 17 intervals, so the automatic labels undercount narration. A positive pair is therefore an audio segment and the scene shown at the same time, with or without the sound source in view.

\paragraph{Filtering and down-weighting.}
The automatic labeler tags all of the about 130K training intervals, of which about 57\% are visible-source. We train three variants of SyncRA on Qwen3-Omni under the Run~1 protocol. \emph{Filter} applies the loss to visible-source intervals only. \emph{Down-weight} gives narration and music intervals a weight of 0.3. \emph{Random filter} removes a random 43\% of the intervals, which matches the interval count of Filter.

\begin{table}[htbp]
\centering\small
\setlength{\tabcolsep}{5pt}
\caption{\textbf{Interval filtering and down-weighting on Qwen3-Omni.} $\Delta$ is the AllFour difference from SyncRA with paired 95\% source-video bootstrap intervals. $P(\mathrm{all}\,|\,\mathrm{O})$ is AllFour divided by original-version accuracy. R@1 is unprojected same-interval retrieval at block 36; values for the three variants use a refreshed pipeline (differs from Table~\ref{tab:representation} by $\le$0.1).}
\label{tab:app_filter}
\begin{tabular}{@{}lrrlrrr@{}}
\toprule
Variant & AllFour & $\Delta$ & 95\% interval & $P(\mathrm{all}\,|\,\mathrm{O})$ & R@1 & Daily-Omni\\
\midrule
SyncRA (all intervals) & 62.50 & \textemdash & \textemdash & 67.02 & 96.9 & 78.95\\
Filter        & 57.25 & $-5.25$ & $[-9.25,-1.25]$ & 61.89 & 93.8 & 78.86\\
Down-weight   & 60.25 & $-2.25$ & $[-6.00,+1.50]$ & 64.61 & 95.9 & 79.03\\
Random filter & 60.75 & $-1.75$ & $[-5.75,+2.25]$ & 65.50 & 95.2 & 78.61\\
\bottomrule
\end{tabular}
\end{table}

\begin{table}[htbp]
\centering\small
\setlength{\tabcolsep}{8pt}
\caption{AllFour (\%) by cue type on the Qwen3-Omni diagnostic. The 400 cues split into 220 with an on-screen speaker and 180 with an off-screen voice.}
\label{tab:app_filter_cues}
\begin{tabular}{@{}lrr@{}}
\toprule
Checkpoint & On-screen-speaker cues (220) & Off-screen-voice cues (180)\\
\midrule
Vanilla SFT            & 37.27 & 30.00\\
SyncRA (all intervals) & 65.91 & 58.33\\
Filter                 & 65.00 & 47.78\\
Down-weight            & 64.55 & 55.00\\
Random filter          & 63.18 & 57.78\\
\bottomrule
\end{tabular}
\end{table}

Filter lowers AllFour by 5.25 points, and its interval excludes zero (Table~\ref{tab:app_filter}). Random filter lowers it by 1.75 points and Down-weight by 2.25 points, and both intervals contain zero. The difference between Filter and Random filter is $-3.50$ $[-7.50,+0.50]$. The split by cue type shows where the drop occurs (Table~\ref{tab:app_filter_cues}). Under Filter, off-screen-voice cues fall from 58.33\% to 47.78\%, which is 10.00 points below Random filter, while on-screen-speaker cues change by less than 1 point. Intervals without a visible source still supply input-time pairs, and these pairs help on off-screen-voice cues, which is consistent with the delay control in Appendix~\ref{app:audio_controls}. Daily-Omni stays between 78.61\% and 79.03\% across the four variants. We therefore keep all intervals as the default.

\clearpage
\section{Full Benchmark Results and Objective Comparisons}
\label{app:full_results}
Tables~\ref{tab:objective-full-0}--\ref{tab:objective-full-3} report complete task-level objective comparisons, Tables~\ref{tab:runs-0}--\ref{tab:runs-3} give the three training runs of Vanilla SFT, SyncRA, and the three auxiliary controls, and Table~\ref{tab:single-evaluations} gives the single Base and Text CoT-SFT evaluations. Tables~\ref{tab:macro-runs-dense} and~\ref{tab:macro-runs-moe} report per-run task-equal macro averages. All comparisons use the fixed input sets in Appendix~\ref{app:benchmark_protocol}.

For task $t$ and run $s$, let $x_{st}=100C_{st}/n_t$. We compute the task mean and sample SD as $\bar x_t=\frac13\sum_s x_{st}$ and $\sqrt{\frac12\sum_s(x_{st}-\bar x_t)^2}$. A $T$-task macro first forms $m_s=T^{-1}\sum_t x_{st}$, then applies the same three-run summary to $m_s$. Differences use unrounded means, with final values rounded to two decimals.
\begin{table}[H]
\centering
\small
\setlength{\tabcolsep}{4.0pt}
\caption{Qwen2.5-Omni-7B: complete training-objective summary. Accuracy in \%, mean $\pm$ sample SD over three independent training runs with different random seeds. Bold marks the highest mean in each row.}
\label{tab:objective-full-0}
\begin{tabular}{@{}lrrrrr@{}}
\toprule
Benchmark & Vanilla SFT & \shortstack{Permuted\\pairs} & \shortstack{Fixed time\\codes} & \shortstack{Clip-level\\AV} & SyncRA \\
\midrule
WorldSense & $50.63\,\pm\,0.32$ & $50.26\,\pm\,0.22$ & $49.73\,\pm\,0.17$ & $50.42\,\pm\,0.16$ & $\mathbf{52.74\,\pm\,0.28}$ \\
Daily-Omni & $68.84\,\pm\,0.36$ & $68.50\,\pm\,0.17$ & $69.53\,\pm\,0.35$ & $69.37\,\pm\,0.26$ & $\mathbf{72.26\,\pm\,0.29}$ \\
OmniVideoBench & $37.77\,\pm\,0.31$ & $38.23\,\pm\,0.15$ & $37.70\,\pm\,0.20$ & $37.33\,\pm\,0.15$ & $\mathbf{38.57\,\pm\,0.25}$ \\
LVOmniBench & $38.03\,\pm\,0.21$ & $39.32\,\pm\,0.21$ & $37.34\,\pm\,0.06$ & $39.51\,\pm\,0.15$ & $\mathbf{39.55\,\pm\,0.30}$ \\
AVUT-Human & $67.97\,\pm\,0.31$ & $69.76\,\pm\,0.44$ & $68.05\,\pm\,0.22$ & $69.28\,\pm\,0.23$ & $\mathbf{71.46\,\pm\,0.34}$ \\
\addlinespace[2pt]
Avg-5 & $52.65\,\pm\,0.12$ & $53.22\,\pm\,0.15$ & $52.47\,\pm\,0.09$ & $53.18\,\pm\,0.12$ & $\mathbf{54.92\,\pm\,0.11}$ \\
Avg-3 & $62.48\,\pm\,0.23$ & $62.84\,\pm\,0.20$ & $62.44\,\pm\,0.20$ & $63.02\,\pm\,0.20$ & $\mathbf{65.49\,\pm\,0.21}$ \\
\bottomrule
\end{tabular}
\end{table}

\begin{table}[H]
\centering
\small
\setlength{\tabcolsep}{4.0pt}
\caption{MiniCPM-o 4.5: complete training-objective summary. Accuracy in \%, mean $\pm$ sample SD over three independent training runs with different random seeds. Bold marks the highest mean in each row.}
\label{tab:objective-full-1}
\begin{tabular}{@{}lrrrrr@{}}
\toprule
Benchmark & Vanilla SFT & \shortstack{Permuted\\pairs} & \shortstack{Fixed time\\codes} & \shortstack{Clip-level\\AV} & SyncRA \\
\midrule
WorldSense & $56.78\,\pm\,0.19$ & $51.53\,\pm\,0.18$ & $52.39\,\pm\,0.08$ & $52.36\,\pm\,0.14$ & $\mathbf{57.61\,\pm\,0.13}$ \\
Daily-Omni & $79.67\,\pm\,0.29$ & $78.08\,\pm\,0.34$ & $77.75\,\pm\,0.10$ & $77.08\,\pm\,0.38$ & $\mathbf{80.51\,\pm\,0.21}$ \\
OmniVideoBench & $35.53\,\pm\,0.32$ & $37.10\,\pm\,0.17$ & $35.80\,\pm\,0.30$ & $36.83\,\pm\,0.12$ & $\mathbf{37.40\,\pm\,0.26}$ \\
LVOmniBench & $31.72\,\pm\,0.21$ & $32.15\,\pm\,0.26$ & $\mathbf{33.53\,\pm\,0.10}$ & $30.24\,\pm\,0.06$ & $31.95\,\pm\,0.26$ \\
AVUT-Human & $78.27\,\pm\,0.23$ & $77.17\,\pm\,0.03$ & $77.15\,\pm\,0.23$ & $76.50\,\pm\,0.03$ & $\mathbf{79.28\,\pm\,0.21}$ \\
\addlinespace[2pt]
Avg-5 & $56.39\,\pm\,0.18$ & $55.21\,\pm\,0.02$ & $55.32\,\pm\,0.00$ & $54.60\,\pm\,0.11$ & $\mathbf{57.35\,\pm\,0.15}$ \\
Avg-3 & $71.57\,\pm\,0.24$ & $68.93\,\pm\,0.06$ & $69.09\,\pm\,0.07$ & $68.65\,\pm\,0.16$ & $\mathbf{72.47\,\pm\,0.17}$ \\
\bottomrule
\end{tabular}
\end{table}

\begin{table}[H]
\centering
\small
\setlength{\tabcolsep}{4.0pt}
\caption{Qwen3-Omni-30B-A3B: complete training-objective summary. Accuracy in \%, mean $\pm$ sample SD over three independent training runs with different random seeds. Bold marks the highest mean in each row.}
\label{tab:objective-full-2}
\begin{tabular}{@{}lrrrrr@{}}
\toprule
Benchmark & Vanilla SFT & \shortstack{Permuted\\pairs} & \shortstack{Fixed time\\codes} & \shortstack{Clip-level\\AV} & SyncRA \\
\midrule
WorldSense & $56.26\,\pm\,0.21$ & $57.28\,\pm\,0.08$ & $56.88\,\pm\,0.12$ & $56.95\,\pm\,0.10$ & $\mathbf{57.36\,\pm\,0.13}$ \\
Daily-Omni & $76.16\,\pm\,0.27$ & $78.72\,\pm\,0.17$ & $78.47\,\pm\,0.21$ & $78.84\,\pm\,0.13$ & $\mathbf{79.09\,\pm\,0.13}$ \\
OmniVideoBench & $46.07\,\pm\,0.32$ & $\mathbf{47.17\,\pm\,0.15}$ & $46.37\,\pm\,0.25$ & $46.83\,\pm\,0.15$ & $47.07\,\pm\,0.25$ \\
LVOmniBench & $40.89\,\pm\,0.23$ & $41.95\,\pm\,0.25$ & $40.76\,\pm\,0.11$ & $42.93\,\pm\,0.06$ & $\mathbf{43.92\,\pm\,0.21}$ \\
AVUT-Human & $78.09\,\pm\,0.27$ & $79.11\,\pm\,0.25$ & $79.48\,\pm\,0.12$ & $79.00\,\pm\,0.10$ & $\mathbf{79.71\,\pm\,0.23}$ \\
\addlinespace[2pt]
Avg-5 & $59.50\,\pm\,0.14$ & $60.85\,\pm\,0.06$ & $60.39\,\pm\,0.14$ & $60.91\,\pm\,0.06$ & $\mathbf{61.43\,\pm\,0.08}$ \\
Avg-3 & $70.17\,\pm\,0.12$ & $71.71\,\pm\,0.08$ & $71.61\,\pm\,0.12$ & $71.59\,\pm\,0.03$ & $\mathbf{72.05\,\pm\,0.10}$ \\
\bottomrule
\end{tabular}
\end{table}

\begin{table}[H]
\centering
\small
\setlength{\tabcolsep}{4.0pt}
\caption{Nemotron-3-Nano-Omni-30B-A3B: complete training-objective summary. Accuracy in \%, mean $\pm$ sample SD over three independent training runs with different random seeds. Bold marks the highest mean in each row.}
\label{tab:objective-full-3}
\begin{tabular}{@{}lrrrrr@{}}
\toprule
Benchmark & Vanilla SFT & \shortstack{Permuted\\pairs} & \shortstack{Fixed time\\codes} & \shortstack{Clip-level\\AV} & SyncRA \\
\midrule
WorldSense & $54.32\,\pm\,0.22$ & $55.43\,\pm\,0.05$ & $\mathbf{55.72\,\pm\,0.07}$ & $55.54\,\pm\,0.27$ & $55.65\,\pm\,0.21$ \\
Daily-Omni & $75.80\,\pm\,0.26$ & $76.33\,\pm\,0.38$ & $77.44\,\pm\,0.14$ & $76.50\,\pm\,0.05$ & $\mathbf{78.42\,\pm\,0.27}$ \\
OmniVideoBench & $43.47\,\pm\,0.15$ & $44.20\,\pm\,0.10$ & $44.10\,\pm\,0.10$ & $43.93\,\pm\,0.15$ & $\mathbf{44.93\,\pm\,0.15}$ \\
LVOmniBench & $41.58\,\pm\,0.15$ & $41.52\,\pm\,0.10$ & $41.39\,\pm\,0.15$ & $42.31\,\pm\,0.10$ & $\mathbf{42.93\,\pm\,0.15}$ \\
AVUT-Human & $73.88\,\pm\,0.30$ & $75.26\,\pm\,0.22$ & $75.19\,\pm\,0.12$ & $75.30\,\pm\,0.06$ & $\mathbf{77.67\,\pm\,0.31}$ \\
\addlinespace[2pt]
Avg-5 & $57.81\,\pm\,0.09$ & $58.55\,\pm\,0.12$ & $58.77\,\pm\,0.05$ & $58.72\,\pm\,0.04$ & $\mathbf{59.92\,\pm\,0.11}$ \\
Avg-3 & $68.00\,\pm\,0.09$ & $69.01\,\pm\,0.15$ & $69.45\,\pm\,0.02$ & $69.11\,\pm\,0.12$ & $\mathbf{70.58\,\pm\,0.09}$ \\
\bottomrule
\end{tabular}
\end{table}

\begin{table}[H]
\centering
\small
\setlength{\tabcolsep}{4.3pt}
\caption{Qwen2.5-Omni-7B: three independent training runs with different random seeds. Each run cell gives correct count $C/n$ and accuracy in \%; means and sample SDs use unrounded accuracies. Run numbers are consistent across tasks.}
\label{tab:runs-0}
\begin{tabular}{@{}lrrrr@{}}
\toprule
Objective & \shortstack{Run 1\\$C/n$ (\%)} & \shortstack{Run 2\\$C/n$ (\%)} & \shortstack{Run 3\\$C/n$ (\%)} & Mean $\pm$ SD \\
\midrule
\multicolumn{5}{@{}l}{\textit{WorldSense ($n=3,172$)}} \\
Vanilla SFT & 1,595/3,172 (50.28) & 1,608/3,172 (50.69) & 1,615/3,172 (50.91) & $50.63\,\pm\,0.32$ \\
SyncRA & 1,670/3,172 (52.65) & 1,683/3,172 (53.06) & 1,666/3,172 (52.52) & $52.74\,\pm\,0.28$ \\
Permuted pairs & 1,593/3,172 (50.22) & 1,588/3,172 (50.06) & 1,602/3,172 (50.50) & $50.26\,\pm\,0.22$ \\
Fixed time codes & 1,583/3,172 (49.91) & 1,577/3,172 (49.72) & 1,572/3,172 (49.56) & $49.73\,\pm\,0.17$ \\
Clip-level AV & 1,595/3,172 (50.28) & 1,605/3,172 (50.60) & 1,598/3,172 (50.38) & $50.42\,\pm\,0.16$ \\
\addlinespace[4pt]
\multicolumn{5}{@{}l}{\textit{Daily-Omni ($n=1,197$)}} \\
Vanilla SFT & 826/1,197 (69.01) & 819/1,197 (68.42) & 827/1,197 (69.09) & $68.84\,\pm\,0.36$ \\
SyncRA & 869/1,197 (72.60) & 863/1,197 (72.10) & 863/1,197 (72.10) & $72.26\,\pm\,0.29$ \\
Permuted pairs & 818/1,197 (68.34) & 822/1,197 (68.67) & 820/1,197 (68.50) & $68.50\,\pm\,0.17$ \\
Fixed time codes & 837/1,197 (69.92) & 829/1,197 (69.26) & 831/1,197 (69.42) & $69.53\,\pm\,0.35$ \\
Clip-level AV & 831/1,197 (69.42) & 833/1,197 (69.59) & 827/1,197 (69.09) & $69.37\,\pm\,0.26$ \\
\addlinespace[4pt]
\multicolumn{5}{@{}l}{\textit{OmniVideoBench ($n=1,000$)}} \\
Vanilla SFT & 375/1,000 (37.50) & 381/1,000 (38.10) & 377/1,000 (37.70) & $37.77\,\pm\,0.31$ \\
SyncRA & 383/1,000 (38.30) & 386/1,000 (38.60) & 388/1,000 (38.80) & $38.57\,\pm\,0.25$ \\
Permuted pairs & 382/1,000 (38.20) & 381/1,000 (38.10) & 384/1,000 (38.40) & $38.23\,\pm\,0.15$ \\
Fixed time codes & 375/1,000 (37.50) & 379/1,000 (37.90) & 377/1,000 (37.70) & $37.70\,\pm\,0.20$ \\
Clip-level AV & 373/1,000 (37.30) & 375/1,000 (37.50) & 372/1,000 (37.20) & $37.33\,\pm\,0.15$ \\
\addlinespace[4pt]
\multicolumn{5}{@{}l}{\textit{LVOmniBench ($n=1,014$)}} \\
Vanilla SFT & 388/1,014 (38.26) & 384/1,014 (37.87) & 385/1,014 (37.97) & $38.03\,\pm\,0.21$ \\
SyncRA & 404/1,014 (39.84) & 398/1,014 (39.25) & 401/1,014 (39.55) & $39.55\,\pm\,0.30$ \\
Permuted pairs & 401/1,014 (39.55) & 397/1,014 (39.15) & 398/1,014 (39.25) & $39.32\,\pm\,0.21$ \\
Fixed time codes & 379/1,014 (37.38) & 379/1,014 (37.38) & 378/1,014 (37.28) & $37.34\,\pm\,0.06$ \\
Clip-level AV & 401/1,014 (39.55) & 399/1,014 (39.35) & 402/1,014 (39.64) & $39.51\,\pm\,0.15$ \\
\addlinespace[4pt]
\multicolumn{5}{@{}l}{\textit{AVUT-Human ($n=1,733$)}} \\
Vanilla SFT & 1,172/1,733 (67.63) & 1,180/1,733 (68.09) & 1,182/1,733 (68.21) & $67.97\,\pm\,0.31$ \\
SyncRA & 1,245/1,733 (71.84) & 1,236/1,733 (71.32) & 1,234/1,733 (71.21) & $71.46\,\pm\,0.34$ \\
Permuted pairs & 1,208/1,733 (69.71) & 1,202/1,733 (69.36) & 1,217/1,733 (70.23) & $69.76\,\pm\,0.44$ \\
Fixed time codes & 1,181/1,733 (68.15) & 1,175/1,733 (67.80) & 1,182/1,733 (68.21) & $68.05\,\pm\,0.22$ \\
Clip-level AV & 1,200/1,733 (69.24) & 1,205/1,733 (69.53) & 1,197/1,733 (69.07) & $69.28\,\pm\,0.23$ \\
\bottomrule
\end{tabular}
\end{table}

\begin{table}[H]
\centering
\small
\setlength{\tabcolsep}{4.3pt}
\caption{MiniCPM-o 4.5: three independent training runs with different random seeds. Each run cell gives correct count $C/n$ and accuracy in \%; means and sample SDs use unrounded accuracies. Run numbers are consistent across tasks.}
\label{tab:runs-1}
\begin{tabular}{@{}lrrrr@{}}
\toprule
Objective & \shortstack{Run 1\\$C/n$ (\%)} & \shortstack{Run 2\\$C/n$ (\%)} & \shortstack{Run 3\\$C/n$ (\%)} & Mean $\pm$ SD \\
\midrule
\multicolumn{5}{@{}l}{\textit{WorldSense ($n=3,172$)}} \\
Vanilla SFT & 1,795/3,172 (56.59) & 1,807/3,172 (56.97) & 1,801/3,172 (56.78) & $56.78\,\pm\,0.19$ \\
SyncRA & 1,826/3,172 (57.57) & 1,824/3,172 (57.50) & 1,832/3,172 (57.76) & $57.61\,\pm\,0.13$ \\
Permuted pairs & 1,637/3,172 (51.61) & 1,628/3,172 (51.32) & 1,639/3,172 (51.67) & $51.53\,\pm\,0.18$ \\
Fixed time codes & 1,659/3,172 (52.30) & 1,664/3,172 (52.46) & 1,662/3,172 (52.40) & $52.39\,\pm\,0.08$ \\
Clip-level AV & 1,660/3,172 (52.33) & 1,657/3,172 (52.24) & 1,666/3,172 (52.52) & $52.36\,\pm\,0.14$ \\
\addlinespace[4pt]
\multicolumn{5}{@{}l}{\textit{Daily-Omni ($n=1,197$)}} \\
Vanilla SFT & 950/1,197 (79.37) & 957/1,197 (79.95) & 954/1,197 (79.70) & $79.67\,\pm\,0.29$ \\
SyncRA & 961/1,197 (80.28) & 964/1,197 (80.53) & 966/1,197 (80.70) & $80.51\,\pm\,0.21$ \\
Permuted pairs & 934/1,197 (78.03) & 939/1,197 (78.45) & 931/1,197 (77.78) & $78.08\,\pm\,0.34$ \\
Fixed time codes & 932/1,197 (77.86) & 930/1,197 (77.69) & 930/1,197 (77.69) & $77.75\,\pm\,0.10$ \\
Clip-level AV & 923/1,197 (77.11) & 918/1,197 (76.69) & 927/1,197 (77.44) & $77.08\,\pm\,0.38$ \\
\addlinespace[4pt]
\multicolumn{5}{@{}l}{\textit{OmniVideoBench ($n=1,000$)}} \\
Vanilla SFT & 354/1,000 (35.40) & 353/1,000 (35.30) & 359/1,000 (35.90) & $35.53\,\pm\,0.32$ \\
SyncRA & 371/1,000 (37.10) & 375/1,000 (37.50) & 376/1,000 (37.60) & $37.40\,\pm\,0.26$ \\
Permuted pairs & 369/1,000 (36.90) & 372/1,000 (37.20) & 372/1,000 (37.20) & $37.10\,\pm\,0.17$ \\
Fixed time codes & 358/1,000 (35.80) & 361/1,000 (36.10) & 355/1,000 (35.50) & $35.80\,\pm\,0.30$ \\
Clip-level AV & 367/1,000 (36.70) & 369/1,000 (36.90) & 369/1,000 (36.90) & $36.83\,\pm\,0.12$ \\
\addlinespace[4pt]
\multicolumn{5}{@{}l}{\textit{LVOmniBench ($n=1,014$)}} \\
Vanilla SFT & 320/1,014 (31.56) & 324/1,014 (31.95) & 321/1,014 (31.66) & $31.72\,\pm\,0.21$ \\
SyncRA & 326/1,014 (32.15) & 321/1,014 (31.66) & 325/1,014 (32.05) & $31.95\,\pm\,0.26$ \\
Permuted pairs & 327/1,014 (32.25) & 323/1,014 (31.85) & 328/1,014 (32.35) & $32.15\,\pm\,0.26$ \\
Fixed time codes & 340/1,014 (33.53) & 339/1,014 (33.43) & 341/1,014 (33.63) & $33.53\,\pm\,0.10$ \\
Clip-level AV & 307/1,014 (30.28) & 306/1,014 (30.18) & 307/1,014 (30.28) & $30.24\,\pm\,0.06$ \\
\addlinespace[4pt]
\multicolumn{5}{@{}l}{\textit{AVUT-Human ($n=1,733$)}} \\
Vanilla SFT & 1,352/1,733 (78.02) & 1,360/1,733 (78.48) & 1,357/1,733 (78.30) & $78.27\,\pm\,0.23$ \\
SyncRA & 1,371/1,733 (79.11) & 1,373/1,733 (79.23) & 1,378/1,733 (79.52) & $79.28\,\pm\,0.21$ \\
Permuted pairs & 1,337/1,733 (77.15) & 1,338/1,733 (77.21) & 1,337/1,733 (77.15) & $77.17\,\pm\,0.03$ \\
Fixed time codes & 1,337/1,733 (77.15) & 1,333/1,733 (76.92) & 1,341/1,733 (77.38) & $77.15\,\pm\,0.23$ \\
Clip-level AV & 1,326/1,733 (76.51) & 1,326/1,733 (76.51) & 1,325/1,733 (76.46) & $76.50\,\pm\,0.03$ \\
\bottomrule
\end{tabular}
\end{table}

\begin{table}[H]
\centering
\small
\setlength{\tabcolsep}{4.3pt}
\caption{Qwen3-Omni-30B-A3B: three independent training runs with different random seeds. Each run cell gives correct count $C/n$ and accuracy in \%; means and sample SDs use unrounded accuracies. Run numbers are consistent across tasks.}
\label{tab:runs-2}
\begin{tabular}{@{}lrrrr@{}}
\toprule
Objective & \shortstack{Run 1\\$C/n$ (\%)} & \shortstack{Run 2\\$C/n$ (\%)} & \shortstack{Run 3\\$C/n$ (\%)} & Mean $\pm$ SD \\
\midrule
\multicolumn{5}{@{}l}{\textit{WorldSense ($n=3,172$)}} \\
Vanilla SFT & 1,777/3,172 (56.02) & 1,789/3,172 (56.40) & 1,788/3,172 (56.37) & $56.26\,\pm\,0.21$ \\
SyncRA & 1,815/3,172 (57.22) & 1,820/3,172 (57.38) & 1,823/3,172 (57.47) & $57.36\,\pm\,0.13$ \\
Permuted pairs & 1,820/3,172 (57.38) & 1,816/3,172 (57.25) & 1,815/3,172 (57.22) & $57.28\,\pm\,0.08$ \\
Fixed time codes & 1,806/3,172 (56.94) & 1,800/3,172 (56.75) & 1,807/3,172 (56.97) & $56.88\,\pm\,0.12$ \\
Clip-level AV & 1,810/3,172 (57.06) & 1,804/3,172 (56.87) & 1,805/3,172 (56.90) & $56.95\,\pm\,0.10$ \\
\addlinespace[4pt]
\multicolumn{5}{@{}l}{\textit{Daily-Omni ($n=1,197$)}} \\
Vanilla SFT & 908/1,197 (75.86) & 913/1,197 (76.27) & 914/1,197 (76.36) & $76.16\,\pm\,0.27$ \\
SyncRA & 945/1,197 (78.95) & 948/1,197 (79.20) & 947/1,197 (79.11) & $79.09\,\pm\,0.13$ \\
Permuted pairs & 943/1,197 (78.78) & 940/1,197 (78.53) & 944/1,197 (78.86) & $78.72\,\pm\,0.17$ \\
Fixed time codes & 939/1,197 (78.45) & 937/1,197 (78.28) & 942/1,197 (78.70) & $78.47\,\pm\,0.21$ \\
Clip-level AV & 942/1,197 (78.70) & 944/1,197 (78.86) & 945/1,197 (78.95) & $78.84\,\pm\,0.13$ \\
\addlinespace[4pt]
\multicolumn{5}{@{}l}{\textit{OmniVideoBench ($n=1,000$)}} \\
Vanilla SFT & 457/1,000 (45.70) & 462/1,000 (46.20) & 463/1,000 (46.30) & $46.07\,\pm\,0.32$ \\
SyncRA & 471/1,000 (47.10) & 468/1,000 (46.80) & 473/1,000 (47.30) & $47.07\,\pm\,0.25$ \\
Permuted pairs & 473/1,000 (47.30) & 472/1,000 (47.20) & 470/1,000 (47.00) & $47.17\,\pm\,0.15$ \\
Fixed time codes & 464/1,000 (46.40) & 461/1,000 (46.10) & 466/1,000 (46.60) & $46.37\,\pm\,0.25$ \\
Clip-level AV & 470/1,000 (47.00) & 467/1,000 (46.70) & 468/1,000 (46.80) & $46.83\,\pm\,0.15$ \\
\addlinespace[4pt]
\multicolumn{5}{@{}l}{\textit{LVOmniBench ($n=1,014$)}} \\
Vanilla SFT & 416/1,014 (41.03) & 412/1,014 (40.63) & 416/1,014 (41.03) & $40.89\,\pm\,0.23$ \\
SyncRA & 443/1,014 (43.69) & 447/1,014 (44.08) & 446/1,014 (43.98) & $43.92\,\pm\,0.21$ \\
Permuted pairs & 425/1,014 (41.91) & 423/1,014 (41.72) & 428/1,014 (42.21) & $41.95\,\pm\,0.25$ \\
Fixed time codes & 414/1,014 (40.83) & 412/1,014 (40.63) & 414/1,014 (40.83) & $40.76\,\pm\,0.11$ \\
Clip-level AV & 436/1,014 (43.00) & 435/1,014 (42.90) & 435/1,014 (42.90) & $42.93\,\pm\,0.06$ \\
\addlinespace[4pt]
\multicolumn{5}{@{}l}{\textit{AVUT-Human ($n=1,733$)}} \\
Vanilla SFT & 1,357/1,733 (78.30) & 1,348/1,733 (77.78) & 1,355/1,733 (78.19) & $78.09\,\pm\,0.27$ \\
SyncRA & 1,382/1,733 (79.75) & 1,385/1,733 (79.92) & 1,377/1,733 (79.46) & $79.71\,\pm\,0.23$ \\
Permuted pairs & 1,373/1,733 (79.23) & 1,374/1,733 (79.28) & 1,366/1,733 (78.82) & $79.11\,\pm\,0.25$ \\
Fixed time codes & 1,375/1,733 (79.34) & 1,378/1,733 (79.52) & 1,379/1,733 (79.57) & $79.48\,\pm\,0.12$ \\
Clip-level AV & 1,371/1,733 (79.11) & 1,368/1,733 (78.94) & 1,368/1,733 (78.94) & $79.00\,\pm\,0.10$ \\
\bottomrule
\end{tabular}
\end{table}

\begin{table}[H]
\centering
\small
\setlength{\tabcolsep}{4.3pt}
\caption{Nemotron-3-Nano-Omni-30B-A3B: three independent training runs with different random seeds. Each run cell gives correct count $C/n$ and accuracy in \%; means and sample SDs use unrounded accuracies. Run numbers are consistent across tasks.}
\label{tab:runs-3}
\begin{tabular}{@{}lrrrr@{}}
\toprule
Objective & \shortstack{Run 1\\$C/n$ (\%)} & \shortstack{Run 2\\$C/n$ (\%)} & \shortstack{Run 3\\$C/n$ (\%)} & Mean $\pm$ SD \\
\midrule
\multicolumn{5}{@{}l}{\textit{WorldSense ($n=3,172$)}} \\
Vanilla SFT & 1,731/3,172 (54.57) & 1,720/3,172 (54.22) & 1,718/3,172 (54.16) & $54.32\,\pm\,0.22$ \\
SyncRA & 1,760/3,172 (55.49) & 1,773/3,172 (55.90) & 1,763/3,172 (55.58) & $55.65\,\pm\,0.21$ \\
Permuted pairs & 1,757/3,172 (55.39) & 1,760/3,172 (55.49) & 1,758/3,172 (55.42) & $55.43\,\pm\,0.05$ \\
Fixed time codes & 1,768/3,172 (55.74) & 1,765/3,172 (55.64) & 1,769/3,172 (55.77) & $55.72\,\pm\,0.07$ \\
Clip-level AV & 1,765/3,172 (55.64) & 1,752/3,172 (55.23) & 1,768/3,172 (55.74) & $55.54\,\pm\,0.27$ \\
\addlinespace[4pt]
\multicolumn{5}{@{}l}{\textit{Daily-Omni ($n=1,197$)}} \\
Vanilla SFT & 904/1,197 (75.52) & 910/1,197 (76.02) & 908/1,197 (75.86) & $75.80\,\pm\,0.26$ \\
SyncRA & 940/1,197 (78.53) & 935/1,197 (78.11) & 941/1,197 (78.61) & $78.42\,\pm\,0.27$ \\
Permuted pairs & 914/1,197 (76.36) & 918/1,197 (76.69) & 909/1,197 (75.94) & $76.33\,\pm\,0.38$ \\
Fixed time codes & 929/1,197 (77.61) & 926/1,197 (77.36) & 926/1,197 (77.36) & $77.44\,\pm\,0.14$ \\
Clip-level AV & 916/1,197 (76.52) & 915/1,197 (76.44) & 916/1,197 (76.52) & $76.50\,\pm\,0.05$ \\
\addlinespace[4pt]
\multicolumn{5}{@{}l}{\textit{OmniVideoBench ($n=1,000$)}} \\
Vanilla SFT & 436/1,000 (43.60) & 433/1,000 (43.30) & 435/1,000 (43.50) & $43.47\,\pm\,0.15$ \\
SyncRA & 448/1,000 (44.80) & 451/1,000 (45.10) & 449/1,000 (44.90) & $44.93\,\pm\,0.15$ \\
Permuted pairs & 441/1,000 (44.10) & 443/1,000 (44.30) & 442/1,000 (44.20) & $44.20\,\pm\,0.10$ \\
Fixed time codes & 442/1,000 (44.20) & 440/1,000 (44.00) & 441/1,000 (44.10) & $44.10\,\pm\,0.10$ \\
Clip-level AV & 438/1,000 (43.80) & 441/1,000 (44.10) & 439/1,000 (43.90) & $43.93\,\pm\,0.15$ \\
\addlinespace[4pt]
\multicolumn{5}{@{}l}{\textit{LVOmniBench ($n=1,014$)}} \\
Vanilla SFT & 422/1,014 (41.62) & 420/1,014 (41.42) & 423/1,014 (41.72) & $41.58\,\pm\,0.15$ \\
SyncRA & 435/1,014 (42.90) & 437/1,014 (43.10) & 434/1,014 (42.80) & $42.93\,\pm\,0.15$ \\
Permuted pairs & 420/1,014 (41.42) & 422/1,014 (41.62) & 421/1,014 (41.52) & $41.52\,\pm\,0.10$ \\
Fixed time codes & 420/1,014 (41.42) & 418/1,014 (41.22) & 421/1,014 (41.52) & $41.39\,\pm\,0.15$ \\
Clip-level AV & 428/1,014 (42.21) & 430/1,014 (42.41) & 429/1,014 (42.31) & $42.31\,\pm\,0.10$ \\
\addlinespace[4pt]
\multicolumn{5}{@{}l}{\textit{AVUT-Human ($n=1,733$)}} \\
Vanilla SFT & 1,286/1,733 (74.21) & 1,276/1,733 (73.63) & 1,279/1,733 (73.80) & $73.88\,\pm\,0.30$ \\
SyncRA & 1,342/1,733 (77.44) & 1,352/1,733 (78.02) & 1,344/1,733 (77.55) & $77.67\,\pm\,0.31$ \\
Permuted pairs & 1,300/1,733 (75.01) & 1,306/1,733 (75.36) & 1,307/1,733 (75.42) & $75.26\,\pm\,0.22$ \\
Fixed time codes & 1,301/1,733 (75.07) & 1,305/1,733 (75.30) & 1,303/1,733 (75.19) & $75.19\,\pm\,0.12$ \\
Clip-level AV & 1,306/1,733 (75.36) & 1,304/1,733 (75.25) & 1,305/1,733 (75.30) & $75.30\,\pm\,0.06$ \\
\bottomrule
\end{tabular}
\end{table}

\begin{table}[H]
\centering
\small
\setlength{\tabcolsep}{5.0pt}
\caption{Single evaluations of Base and Text CoT-SFT. Each observed cell gives integer $C/n$ and accuracy in \%. These fixed-checkpoint evaluations have no training-run SD.}
\label{tab:single-evaluations}
\begin{tabular}{@{}llrr@{}}
\toprule
Backbone & Benchmark & Base: $C/n$ (\%) & Text CoT-SFT: $C/n$ (\%) \\
\midrule
Qwen2.5-Omni & WorldSense & 1,436/3,172 (45.27) & 1,520/3,172 (47.92) \\
 & Daily-Omni & 739/1,197 (61.74) & 756/1,197 (63.16) \\
 & OmniVideoBench & 251/1,000 (25.10) & 355/1,000 (35.50) \\
 & LVOmniBench & 286/1,014 (28.21) & 321/1,014 (31.66) \\
 & AVUT-Human & 1,122/1,733 (64.74) & 1,176/1,733 (67.86) \\
\addlinespace[3pt]
MiniCPM-o 4.5 & WorldSense & 1,750/3,172 (55.17) & 1,767/3,172 (55.71) \\
 & Daily-Omni & 953/1,197 (79.62) & 956/1,197 (79.87) \\
 & OmniVideoBench & 278/1,000 (27.80) & 334/1,000 (33.40) \\
 & LVOmniBench & 286/1,014 (28.21) & 320/1,014 (31.56) \\
 & AVUT-Human & 1,353/1,733 (78.07) & 1,362/1,733 (78.59) \\
\addlinespace[3pt]
Qwen3-Omni & WorldSense & 1,745/3,172 (55.01) & 1,757/3,172 (55.39) \\
 & Daily-Omni & 897/1,197 (74.94) & 891/1,197 (74.44) \\
 & OmniVideoBench & 439/1,000 (43.90) & 442/1,000 (44.20) \\
 & LVOmniBench & 405/1,014 (39.94) & 402/1,014 (39.64) \\
 & AVUT-Human & 1,348/1,733 (77.78) & 1,337/1,733 (77.15) \\
\addlinespace[3pt]
Nemotron & WorldSense & 1,677/3,172 (52.87) & 1,678/3,172 (52.90) \\
 & Daily-Omni & 869/1,197 (72.60) & 841/1,197 (70.26) \\
 & OmniVideoBench & 422/1,000 (42.20) & 431/1,000 (43.10) \\
 & LVOmniBench & 397/1,014 (39.15) & 404/1,014 (39.84) \\
 & AVUT-Human & 1,264/1,733 (72.94) & 1,267/1,733 (73.11) \\
\bottomrule
\end{tabular}
\end{table}

\begin{table}[H]
\centering
\small
\setlength{\tabcolsep}{4.4pt}
\caption{Per-run task-equal macro accuracies (\%): dense backbones. Each run is averaged over tasks before computing the across-run mean and sample SD.}
\label{tab:macro-runs-dense}
\begin{tabular}{@{}lllrrrr@{}}
\toprule
Backbone & Macro & Objective & Run 1 & Run 2 & Run 3 & Mean $\pm$ SD \\
\midrule
\addlinespace[3pt]
Qwen2.5-Omni & Avg-5 & Vanilla SFT & 52.54 & 52.63 & 52.78 & $52.65\,\pm\,0.12$ \\
 &  & SyncRA & 55.05 & 54.87 & 54.83 & $54.92\,\pm\,0.11$ \\
 &  & Permuted pairs & 53.20 & 53.07 & 53.38 & $53.22\,\pm\,0.15$ \\
 &  & Fixed time codes & 52.57 & 52.41 & 52.43 & $52.47\,\pm\,0.09$ \\
 &  & Clip-level AV & 53.16 & 53.31 & 53.08 & $53.18\,\pm\,0.12$ \\
\addlinespace[3pt]
 & Avg-3 & Vanilla SFT & 62.31 & 62.40 & 62.74 & $62.48\,\pm\,0.23$ \\
 &  & SyncRA & 65.70 & 65.49 & 65.27 & $65.49\,\pm\,0.21$ \\
 &  & Permuted pairs & 62.75 & 62.70 & 63.08 & $62.84\,\pm\,0.20$ \\
 &  & Fixed time codes & 62.66 & 62.26 & 62.40 & $62.44\,\pm\,0.20$ \\
 &  & Clip-level AV & 62.98 & 63.24 & 62.85 & $63.02\,\pm\,0.20$ \\
\addlinespace[3pt]
MiniCPM-o 4.5 & Avg-5 & Vanilla SFT & 56.19 & 56.53 & 56.47 & $56.39\,\pm\,0.18$ \\
 &  & SyncRA & 57.24 & 57.28 & 57.52 & $57.35\,\pm\,0.15$ \\
 &  & Permuted pairs & 55.19 & 55.21 & 55.23 & $55.21\,\pm\,0.02$ \\
 &  & Fixed time codes & 55.33 & 55.32 & 55.32 & $55.32\,\pm\,0.00$ \\
 &  & Clip-level AV & 54.59 & 54.50 & 54.72 & $54.60\,\pm\,0.11$ \\
\addlinespace[3pt]
 & Avg-3 & Vanilla SFT & 71.32 & 71.80 & 71.59 & $71.57\,\pm\,0.24$ \\
 &  & SyncRA & 72.32 & 72.42 & 72.66 & $72.47\,\pm\,0.17$ \\
 &  & Permuted pairs & 68.93 & 68.99 & 68.87 & $68.93\,\pm\,0.06$ \\
 &  & Fixed time codes & 69.10 & 69.02 & 69.16 & $69.09\,\pm\,0.07$ \\
 &  & Clip-level AV & 68.65 & 68.48 & 68.81 & $68.65\,\pm\,0.16$ \\
\bottomrule
\end{tabular}
\end{table}

\begin{table}[H]
\centering
\small
\setlength{\tabcolsep}{4.4pt}
\caption{Per-run task-equal macro accuracies (\%): mixture-of-experts backbones. Each run is averaged over tasks before computing the across-run mean and sample SD.}
\label{tab:macro-runs-moe}
\begin{tabular}{@{}lllrrrr@{}}
\toprule
Backbone & Macro & Objective & Run 1 & Run 2 & Run 3 & Mean $\pm$ SD \\
\midrule
\addlinespace[3pt]
Qwen3-Omni & Avg-5 & Vanilla SFT & 59.38 & 59.46 & 59.65 & $59.50\,\pm\,0.14$ \\
 &  & SyncRA & 61.34 & 61.48 & 61.47 & $61.43\,\pm\,0.08$ \\
 &  & Permuted pairs & 60.92 & 60.80 & 60.82 & $60.85\,\pm\,0.06$ \\
 &  & Fixed time codes & 60.39 & 60.25 & 60.53 & $60.39\,\pm\,0.14$ \\
 &  & Clip-level AV & 60.97 & 60.85 & 60.90 & $60.91\,\pm\,0.06$ \\
\addlinespace[3pt]
 & Avg-3 & Vanilla SFT & 70.06 & 70.15 & 70.30 & $70.17\,\pm\,0.12$ \\
 &  & SyncRA & 71.97 & 72.16 & 72.01 & $72.05\,\pm\,0.10$ \\
 &  & Permuted pairs & 71.79 & 71.69 & 71.64 & $71.71\,\pm\,0.08$ \\
 &  & Fixed time codes & 71.57 & 71.51 & 71.75 & $71.61\,\pm\,0.12$ \\
 &  & Clip-level AV & 71.62 & 71.56 & 71.60 & $71.59\,\pm\,0.03$ \\
\addlinespace[3pt]
Nemotron & Avg-5 & Vanilla SFT & 57.90 & 57.72 & 57.81 & $57.81\,\pm\,0.09$ \\
 &  & SyncRA & 59.83 & 60.04 & 59.89 & $59.92\,\pm\,0.11$ \\
 &  & Permuted pairs & 58.46 & 58.69 & 58.50 & $58.55\,\pm\,0.12$ \\
 &  & Fixed time codes & 58.81 & 58.71 & 58.79 & $58.77\,\pm\,0.05$ \\
 &  & Clip-level AV & 58.71 & 58.69 & 58.75 & $58.72\,\pm\,0.04$ \\
\addlinespace[3pt]
 & Avg-3 & Vanilla SFT & 68.10 & 67.96 & 67.94 & $68.00\,\pm\,0.09$ \\
 &  & SyncRA & 70.48 & 70.67 & 70.58 & $70.58\,\pm\,0.09$ \\
 &  & Permuted pairs & 68.92 & 69.18 & 68.93 & $69.01\,\pm\,0.15$ \\
 &  & Fixed time codes & 69.47 & 69.44 & 69.44 & $69.45\,\pm\,0.02$ \\
 &  & Clip-level AV & 69.18 & 68.97 & 69.19 & $69.11\,\pm\,0.12$ \\
\bottomrule
\end{tabular}
\end{table}

\end{document}